\documentclass{article}

\usepackage[final]{neurips}

\usepackage[utf8]{inputenc} 
\usepackage[T1]{fontenc}    
\usepackage{hyperref}       
\usepackage{url}            
\usepackage{booktabs}       
\usepackage{amsfonts}       
\usepackage{nicefrac}       
\usepackage{microtype}      
\usepackage{xcolor}         
\usepackage{amsthm}
\usepackage{amsmath}
\usepackage{multirow}
\usepackage{graphicx}
\usepackage{float}
\usepackage{subfig}
\usepackage{makecell}

\usepackage{arydshln}
\usepackage{tcolorbox}
\usepackage{hypcap}
\usepackage{CJKutf8}

\title{MindLLM: Pre-training Lightweight Large Language Model from Scratch, Evaluations and Domain Applications}

\author{%
  Yizhe Yang
   \And
  Huashan Sun
   \AND
  Jiawei Li
  \And
  Runheng Liu
  \And
  Yinghao Li
  \And
  Yuhang Liu
  \And
  Yang Gao
  \And
  Heyan Huang \\
  School of Computer Science\\
  Beijing Institute of Technology\\
  \texttt{author@bit.edu.cn} \\
}

\author{%
Yizhe Yang$^{*}$ \quad Huashan Sun\thanks{~~Equal contribution} \quad Jiawei Li \quad Runheng Liu \quad Yinghao Li \\
\textbf{Yuhang Liu} \quad    \textbf{Yang Gao}\thanks{~~Corresponding author} \quad \textbf{Heyan Huang}$^{\dagger}$\\
School of Computer Science Beijing Institute of Technology\\ 
Beijing Engineering Research Center of High Volume Language Information Processing \\
        and Cloud Computing Applications\\
Beijing Institute of Technology Southeast Academy of Information Technology\\
\texttt{\{yizheyang,hssun,jwli,rhliu,yhli,codelyh,hhy63,gyang\}@bit.edu.cn}
}

\begin{document}

\maketitle

\begin{abstract}
\end{abstract}

 Large Language Models (LLMs) have demonstrated remarkable performance across various natural language tasks, marking significant strides towards general artificial intelligence. While general artificial intelligence is leveraged by developing increasingly large-scale models, there could be another branch to develop lightweight custom models that better serve certain domains, taking into account the high cost of training and deploying LLMs and the scarcity of resources. In this paper, we present MindLLM, a novel series of bilingual lightweight large language models, trained from scratch, alleviating such burdens by offering models with 1.3 billion and 3 billion parameters. A thorough account of experiences accrued during large model development is given, covering every step of the process, including data construction, model architecture, evaluation, and applications. Such insights are hopefully valuable for fellow academics and developers. MindLLM consistently matches or surpasses the performance of other open-source larger models on some public benchmarks. We also introduce an innovative instruction tuning framework tailored for smaller models to enhance their capabilities efficiently. Moreover, we explore the application of MindLLM in specific vertical domains such as law and finance, underscoring the agility and adaptability of our lightweight models.


\section{Introduction}
\label{sec:introduction}

The realm of large language models has seen remarkable advancements in recent years. Models such as BERT~\citep{kenton2019bert} and GPT~\citep{radford2018improving}, once contained millions of parameters, but this number has escalated to billions or even trillions in contemporary models like GPT-3~\citep{brown2020language} and PaLM~\citep{chowdhery2022palm}. This increase in scale has fueled significant enhancements in language models, bestowing upon them a near-human fluency and the ability to adeptly execute diverse natural language tasks. Large Language Models (LLMs) such as ChatGPT~\citep{chatgpt}, GPT-4~\citep{gpt4}, PaLM-2~\citep{anil2023palm}, LLaMA~\citep{touvron2023llama}, Falcon~\citep{penedo2023refinedweb}, Baichuan-2~\citep{yang2023baichuan} among others, showcase high linguistic proficiency across various domains from casual conversation to detailed conceptual elaboration. Their capacity for generating human-like text has significantly piqued public interest. Such a breakthrough highlights the potential of LLMs in general artificial intelligence. However, inefficiencies persist when attempting to incorporate LLMs into specialized domain applications due to resource constraints. 
To our knowledge, domain-specific applications often do not require extensive general knowledge retention or capabilities such as program execution, multi-task functionality, or model calibration, despite LLMs demonstrating vast knowledge retention and emergent capabilities~\citep{wei2022emergent}. Instead, models should be tailored for well-defined tasks and domain-specific knowledge retention. For example, Phi-1, trained on high-quality code data with merely 1.3 billion parameters, significantly outperforms larger models~\citep{gunasekar2023textbooks}. Furthermore, these models should be readily accessible for developers in relevant domains without incurring prohibitive costs or risks related to data leakage.

In this paper, we present a series of lightweight bilingual large language models, \textbf{Min}g\textbf{D}e LLM, short for MindLLM. MindLLM has two separate models, MindLLM-1.3B with 1.3 billion parameters and MindLLM-3B with 3.1 billion parameters. The MindLLM-1.3B is trained on 323 billion tokens while the MindLLM-3B is trained on 500 billion tokens in a mixture of bilingual data. Considering the scalability, learning capacity, and training costs of the model, we have adopted a series of optimization operators such as RoPE~\citep{su2021roformer} and FlashAttention-2~\citep{dao2023flashattention}. We pre-train our models in two distinct training strategies: (1) training on binlingual mixture data and (2) training on monolingual data then transfer. Our experiments show that although training on new language data after monolingual pre-training can implement language transferring and improve capability by language alignment, training from scratch with bilingual data is a better way for capability learning and avoiding catastrophic forgetting~\citep{Continual_Learning_survey}.

Despite its lightweight nature, MindLLMs deliver impressive performance. For general English capabilities on MMLU~\citep{MMLU} and AGIEval~\citep{AGIEVAL}, MindLLM-1.3B (24.81) and MindLLM-3B (24.53) outperform larger counterparts such as MPT-7B~\citep{MosaicML2023Introducing} (23.92), MOSS-Base-16B~\citep{sun2023moss} (22.6), and Bloom-7B~\citep{Bloom} (24.25), achieving results close to Falcon-7B~\citep{refinedweb} (25.05). On the C-Eval~\citep{ceval} and CMMLU~\citep{CMMLU} benchmark, MindLLM-1.3B demonstrates a superior average performance than MindLLM-3B, comparable to Open-LLaMA-7B~\citep{openlm2023openllama} and Bloom-3B. 
Meanwhile, MindLLM-3B performs on par with Bloom-7B and surpasses Open-LLaMA-3B. 
In our specialized experiments to assess unique capabilities, MindLLM-3B exhibits exceptional performance in mathematics and bilingualism. Specifically, MindLLM-3B achieves a score of 44.75 in arithmetic ability, outperforming MPT-7B~\citep{MosaicML2023Introducing} (28.26) and MOSS-Base-16B~\citep{sun2023moss} (37.82) and on par with Falcon-7B~\citep{refinedweb} (45.76), all of which are pre-trained on larger datasets or have larger scale compared to MindLLM. For bilingual capability, our experiment shows that only Blooms and MindLLM-3B exhibit excellent alignment in both directions in Chinese and English, surpassing larger multilingual models like Baichuan2-7B~\citep{baichuan2023baichuan2} and LLaMA2-7B~\citep{DBLP:journals/corr/abs-2307-09288} and bilingual models like MOSS-Base-16B~\citep{sun2023moss}.

Additionally, we execute instruction tuning on our MindLLM and make some intriguing observations. We observe that lightweight models, despite encountering challenges in augmenting their comprehensive competencies through diverse instruction-tuning datasets, can achieve significant advancements in their specific proficiencies when exposed to datasets meticulously tailored to their target domains. In the context of enhancing proficiency in Chinese subject matter, the MindLLM-1.3B model, when tuned with specifically tailored datasets, exhibits superior performance compared to counterparts,  including the ChatGLM-6B~\citep{du2022glm}, Yuren-13B, and Chinese-Alpaca-33B~\citep{chinese-llama-alpaca}, yielding an accuracy enhancement exceeding 15\%. 
When tasked with English reasoning assignments, the MindLLM-1.3B model demonstrates a notable enhancement of approximately 5\% in the domain of tasks concerning integrated and knowledge-based reasoning. The significance of this capability becomes particularly pronounced when considering its application within specialized domains. 
To facilitate this, we introduce the approach to construct an instruction set using an entropy-based quality filtering strategy and illustrate its efficacy in sieving out high-quality instruction tuning data that is well-suited for lightweight models. We have empirically demonstrated that, for lightweight models, data quality is more important than data diversity and quantity. And we provide specific recommendations for filtering high-quality instruction tuning data.

Furthermore, we have applied our MindLLM to two specialized domains: law and finance. Our results demonstrate that our models can achieve comparable performance while requiring lower computational costs. We conduct supervised fine-tuning on MindLLM 1.3B and 3B using legal datasets of various sizes, outperforming other open-source models with equivalent parameter scales and being on par with some models with a parameter size of 7B. Furthermore, we perform empirical experiments on MindLLM models in the domain of finance, employing supervised fine-tuning and chain-of-thought training methods. The conducted experiments validate that harnessing limited auxiliary data for training purposes can significantly unleash the latent potential of the MindLLM models.

In this paper, we offer insights into our detailed training procedures, the trials we conducted, and the valuable lessons we learned throughout the training process. We share our experiences gained from comparing large and lightweight models and our endeavors to apply lightweight models to specific domains. In the forthcoming sections, we will provide a comprehensive overview of our training data, model architecture, and training methodologies. We will then benchmark the performance of our models against other larger language models on a standardized test suite and discuss effective strategies for fine-tuning lightweight models. Finally, we will showcase our attempts to apply our lightweight model to specialized domains.

The highlights of our model can be summarized as follows:

\begin{enumerate}
\item This study presents MindLLM, a novel bilingual lightweight large language model trained from scratch. Diverse bilingual training data is collected and used for pre-training, guided by preliminary experiments on data. Our experience with data handling is also shared, which includes maintaining a high-quality and high-proportion of webtext, retaining long-term data like books and dialogues, downsampling math data while upsampling code data. Based on our experiment concerning data curricula, we recommend evenly shuffling data for capability learning but chunking some samples for few-shot learning scenarios.

\item Our evaluation results show that MindLLMs outperform larger models like MPT-7B and GPT-J-6B on MMLU and AGIEval. As for Chinese general capabilities, MindLLMs exhibit comparable performance to larger counterparts on C-Eval and CMMLU. Specifically, MindLLM-3B outperforms larger models such as MOSS-Base-16B, MPT-7B in mathematical abilities and surpasses Baichuan2-7B and MOSS-Base-16B in bilingual capability. Moreover, MindLLM-1.3B is better than the equally sized GPT-Neo-1.3B in mathematics.
\item We conducted experiments to compare two distinct training strategies in bilingual learning and investigate the impact of whether maintaining a uniform distribution of data during pre-training. Combined with other counterparts' evaluation results, we conclude that it is suboptimal for lightweight models ($\leq$ 7B) with constrained capacity scale to achieve complex capabilities such as mathematics, reasoning, or bilingual alignment through pre-training then transfer-training strategy, as integrating new and existing knowledge proves difficult. In contrast, a more efficacious strategy involves the strategic integration of diverse data types aligned with the requirements of downstream tasks from scratch, thereby ensuring the stable and effective acquisition of the desired capabilities.

\item We find that lightweight models face a big challenge in augmenting their general capabilities with diverse instruction tuning data. Leveraging tailored data for a particular ability during instruction tuning can significantly enhance the specific ability (such as integrated reasoning ability or subject-knowledge ability) of lightweight models.
\item We introduce the approach to construct an instruction set using an entropy-based quality filtering strategy and demonstrate its effectiveness in filtering high-quality instruction tuning data for lightweight models. We prove that, in the context of lightweight models, the optimization of model performance is more proficiently realized through the amelioration of instruction tuning data quality, as opposed to mere augmentation in data quantity. 
\item Our models showcase outstanding performance in specific domains, particularly in areas like law and finance. We find that the difference in model parameter sizes does not yield significant distinctions within specific domains, and smaller models can outperform their larger counterparts. Our models outperform all models with parameter sizes ranging from 1.3B to 3B in specific domains while remaining competitive with models having parameter sizes from 6B to 13B. The model's classification ability within specific domains shows a noticeable enhancement under the COT method.
\end{enumerate}

\section{Data}
\label{sec:data}

This section provides an in-depth exploration of our pre-training data, elucidating the critical aspects of its structure, source, and the practical insights derived during the pre-training phase. Our training corpus is a diverse blend of both English and Chinese language data sources. The English component originates from the Pile dataset~\citep{gao2020pile}, and the Chinese component comprises data from Wudao~\citep{yuan2021wudaocorpora}, CBooks\footnote{https://github.com/FudanNLPLAB/CBook-150K}, and data meticulously gathered through web crawling. 

To ensure data quality, we execute a thorough preprocessing pipeline, which involves purging special tags via rigorous data cleaning, data deduplication using Locality-Sensitive Hashing (LSH), and comprehensive filtering to eliminate low-quality content predominantly from advertisements or inappropriate material. We also examine the relationship between data volume and model capacity, assess the impact of different data types on model fitting effectiveness, and evaluate model training stability when handling mixed data sources. This analysis offers valuable insights into the vital role of pre-training data and the complexities of processing it. An overview of our final processed data can be found in Table~\ref{tab:data overview}, and we apply some mixture craftsmanship to construct training data based on data engineering and experience.

\begin{table}[]
\centering
\begin{tabular}{lrrrr}
\hline
\multirow{2}{*}{\textbf{Category}} & \multicolumn{2}{c}{\textbf{English}} & \multicolumn{2}{c}{\textbf{Chinese}} \\ \cline{2-5} 
                          & \textbf{Store Size}   & \textbf{Token Size}   & \textbf{Store Size}    & \textbf{Token Size}  \\ \hline
Academic                  & 242.17 GiB   & 79.91 B      & 69.49 GiB     & 20.27 B     \\
Book                      & 118.14 GiB   & 29.82 B      & 585.97 GiB    & 150.23 B    \\
Code                      & 95.16 GiB    & 41.98 B      & 15.31 GiB     & 13.17 B     \\
Encyclopedia              & 6.38 GiB     & 1.51 B       & 13.45 GiB     & 4.67 B      \\
Math                      & 7.75 GiB     & 6.31 B       & 14.96 GiB     & 11.74 B     \\
QA                        & 32.20 GiB    & 11.06 B      & 31.13 GiB     & 10.28 B     \\
Webtext                   & 227.12 GiB   & 55.87 B      & 643.32 GiB    & 172.73 B    \\
Dialogue                  & 17.59 GiB    & 7.65 B       & 107.46 GIB    & 31.87 B     \\
Technology                & 28.67 GiB    & 10.49 B      & 43.58 GiB     & 17.64 B     \\ \hline
Total                     & 825.18 GiB   & 244.62 B     & 1524.67 GiB   & 432.6 B     \\ \hline
\end{tabular}
\caption{The overview of our processed data. The final mixed training data is constructed based on data engineer and experience.}
\label{tab:data overview}
\end{table}

\subsection{English Pre-training Dataset} 

We utilize the Pile~\citep{gao2020pile} as our primary source for English pre-training datasets. The Pile, an 800GB dataset derived from 22 diverse sources including web scrapes, academic resources, books, and more, closely mirrors the data sources used in larger non-open source models like PaLM~\citep{chowdhery2022palm}, Chinchilla~\citep{hoffmann2022empirical}, and GPT-3~\citep{brown2020language}. Following previous works~\citep{brown2020language,gao2020pile} and our preliminary experiments, we increase the weights of higher-quality compositions and important categories by up-sampling.

\subsection{Chinese Pre-training Dataset} 

Our Chinese pre-training dataset amalgamates contributions from WuDao~\citep{yuan2021wudaocorpora}, Cook and open-source data from various Chinese websites.

WuDao is culled from 3 billion web pages with high text density. We employ a set of cleaning processes to enhance the quality of the corpus and filter sensitive and personal information. Although the original WuDao corpus is vast (3TB), we only utilize its open version due to its close-source nature. The version used approximates 200GB of training data, encompassing multiple domains like wiki, news, and comments.

CBook, a large-scale Chinese book corpora collected through open-source MD5 book links by FudanNLPLAB, forms an integral part of our training dataset. Recognizing the value of books for long-range context modeling and coherent storytelling, we parse and extract text content from the corpus in various formats such as pdf, mobi, and epub totaling over 100 thousand books.

Given that the previous datasets only comprised roughly 300GB of training data, it is necessary to supplement this with additional data to train a large-scale language model effectively. We therefore gather a substantial Chinese text corpus from various websites. This self-collected data underwent strict processing procedures to structure it in a unified format and filter out non-content tokens and sensitive information. The final dataset contains about 1.8TB of data, spanning diverse domains such as novels, literature, law, and more. The processing details for the raw data are discussed in Section~\ref{sec:data processing}.

\subsection{Data Processing}
\label{sec:data processing}
Webpage text content is typically ample for training our language models within one epoch. However, we've observed that unfiltered web text often exhibits lower quality than more curated datasets and contains sensitive, irrelevant, or inappropriate information. To retain only the informative and diverse data for training, we executed several processing stages.

\paragraph{Format Cleaning} We employ webpage parsers to extract and clean text content from source webpages. This stage involves eliminating unhelpful HTML, CSS, and JS identifiers, along with abnormal symbols like emojis to ensure textual fluidity. Moreover, we deploy a decoding test and high-frequency garbled vocabulary filter to exclude garbled content. We also address formatting inconsistencies such as superfluous spaces separating Chinese characters and converting special invisible characters to spaces. Notably, we preserve traditional Chinese characters to allow our model to learn from ancient literature or poetry, despite recognizing the distinct informational distribution compared to modern texts (refer to~\ref{sec:data experience}).

\paragraph{Low-Quality Content Filtering} We assess data quality based on the text-to-content ratio in webpage content. Specifically, we ignore webpages with text density below 75\% or those containing fewer than 100 Chinese characters. This threshold is determined through preliminary tests on sampled webpages.

\paragraph{Dedumplication} As WuDao also draws from webpages, and some websites repost identical information, we implement Locality-Sensitive Hashing (LSH), particularly the SimHash algorithm, to eliminate duplicated content and conserve our training data's diversity.

\paragraph{Sensitive Information Exclusion} Webpages often harbor sensitive content such as offensive words, inflammatory comments, sexually explicit materials, and illegal contents. To build a positive and responsible language model, we use heuristics and a sensitive word vocabulary to detect and filter this material. For privacy protection, we identify private information like identity numbers, phone numbers, and email addresses using regular expressions, replacing them with special tokens.

\paragraph{Self-Repeating Content Filter} Low-information data, like advertisements, often exhibits repetition. We thus analyze phrase frequency within website text content. While some repetitive phrases are natural and acceptable, such as in news or academic papers, we posit that frequent repeating phrases from a common website may not be desirable for model learning. Therefore, our filter focuses on continuous repetitive phrases typical of advertisements or website information, especially for uncertified websites.

\subsection{Empirical Observations on Data Engineer}
\label{sec:data experience}
\subsubsection{The Relationship between Data Volume and Model Capacity} 

\begin{figure}[htb]
    \centering
    \includegraphics[width=\textwidth]{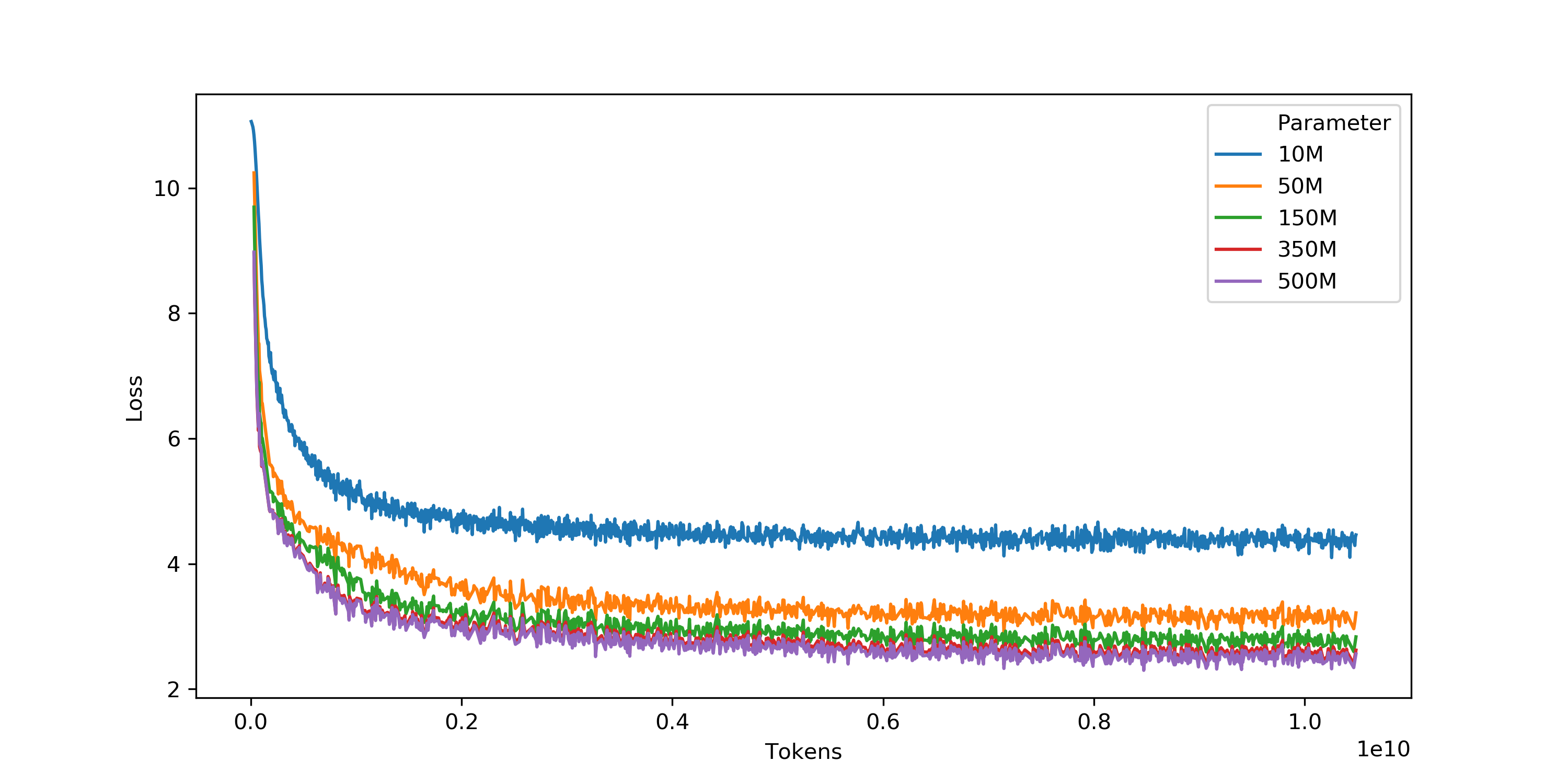}
    \caption{The various training loss of small models for scaling law.}
    \label{fig:small losss}
\end{figure}

To ensure optimal performance in the face of increasing training costs for deep learning and large language models, we investigate the relationship between data volume and model capacity, specifically, the neural scaling laws. These laws depict error reduction as a power function of training data volume, model capacity, or both. Before embarking on the training of large language models with billions of parameters, we initially train smaller models to establish a scaling law for training larger ones. We employ model sizes spanning from 10 million to 500 million parameters, with each model being trained on up to 10 billion tokens. This training utilizes consistent hyperparameters and the identical dataset mentioned earlier. By analyzing the final loss of various models, we can establish a mapping from training FLOPs to the target loss. As shown in Figure~\ref{fig:small losss}, models of different sizes are saturated with different amounts of training data, and the required training data increases with the increase of model size. To conform to the precise data requirement of target model, we utilize the formula provided by~\citet{henighan2020scaling} as follows to fit the model's scaling law.

\begin{equation}
    \mathcal{L}_C=a\times \log C + \mathcal{L}_{\infty}
\end{equation}

where $\mathcal{L}_{\infty}$ represents the irreducible loss, and the $a\times \log C$ corresponds to the reducible loss formulated as a power-law scaling term. Here, $C$ is training FLOPs and the $\mathcal{L}_C$ signifies the final loss of the model under those FLOPs.

\begin{figure}[htb]
    \centering
    \includegraphics[width=\textwidth]{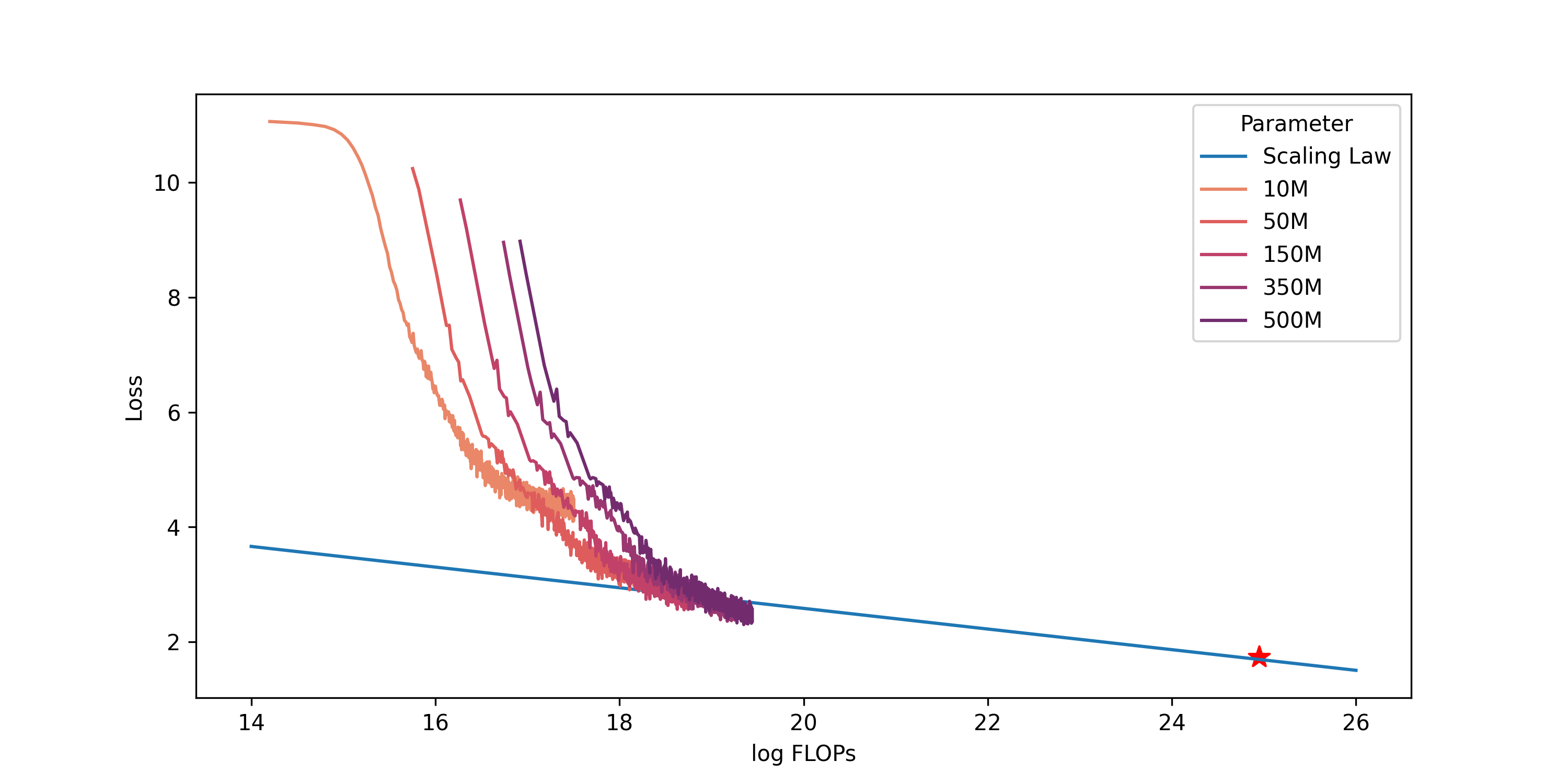}
    \caption{The scaling law of MindLLM. We conduct training on a variety of models, spanning from 10 million to 500 million parameters, utilizing a dataset comprising 10 billion tokens. Through the application of a power-law term to the losses associated with training FLOPs, we make predictions regarding the losses for the training of MindLLM-3B on a dataset of 500 billion tokens. Remarkably, this fitting process accurately predicts the final losses of the models, denoted with {\color{red}$\star$}.}
    \label{fig:scaling law}
\end{figure}

We employ the \texttt{curve\_fit} function from the SciPy\footnote{https://scipy.org} library to determine the parameters. The resulting fits scaling curve is characterized by parameters: $a=0.214, \mathcal{L}_{\infty}=1.692$. Additionally, Figure~\ref{fig:scaling law} displays the final loss prediction for a model with 3 billion parameters.

\subsubsection{The Influence from Data on Model Capability}

Data influence encompasses two critical aspects: (1) Mix Ratio, which pertains to how data from different sources should be combined to create a fixed-size dataset within the constraints of a limited training budget, and (2) Data Curriculum, which addresses the scheduling of data from various sources to instruct the model for specific capabilities.

The mix ratio refers to how data from different sources is weighted and is crucial for the model's ability to learn specific capabilities from the training data. For instance, LLaMA ~\citep{touvron2023llama} downsamples both code and academic data, resulting in relatively lower coding and reasoning performance compared to models trained on more specific data, such as StarCoder~\citep{li2023starcoder} and Galactica~\citep{taylor2022galactica}. \citet{xie2023doremi}  demonstrate that varying the mix ratio can lead to differences in the speed of learning, and an optimized mix ratio for the same dataset enhances model performance with a steeper learning curve, accelerating model learning. Although we do not implement the method in ~\citet{xie2023doremi} to improve our mix ratio (as it is reported after we had completed the data processing), we conduct experiments with smaller models to investigate data mix ratios similar to those used in DoReMi.

\begin{figure}
    \centering
    \includegraphics[width=\textwidth]{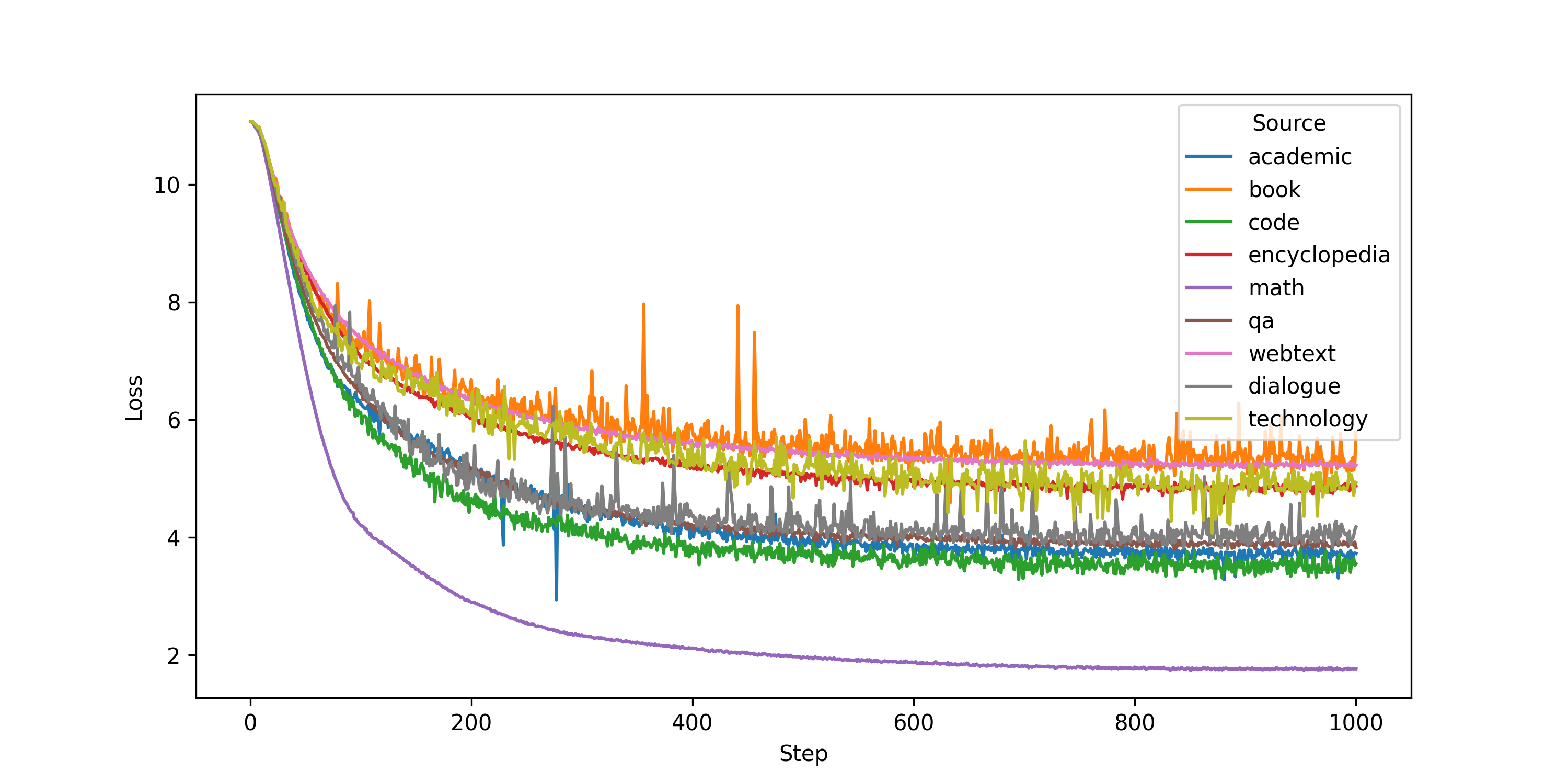}
    \caption{The loss of 15M model trained on data from different sources. Different loss tendence and final loss demonstrate the various characteristics of data.}
    \label{fig:types loss}
\end{figure}

We downsize the data from each source maintaining the relative proportions and train smaller models with 15 million parameters using a single data source. We maintain a consistent number of tokens for model training instead of one-epoch training, as obtaining data from some sources is unfeasible during scaling. As depicted in Figure~\ref{fig:types loss}, different types of data have varying impacts on learning efficiency and the model's final outcomes. For instance, the final loss of math data is much lower with more rapid learning speed, indicating it exhibits a more conspicuous pattern and is easy to learn. In contrast, data from information-rich books or diverse webtext requires slower adaptation. Some domain-similar data may be closed in loss, such as Technology and Encyclopedia. 

\begin{figure}
    \centering
    \includegraphics[width=0.8\textwidth]{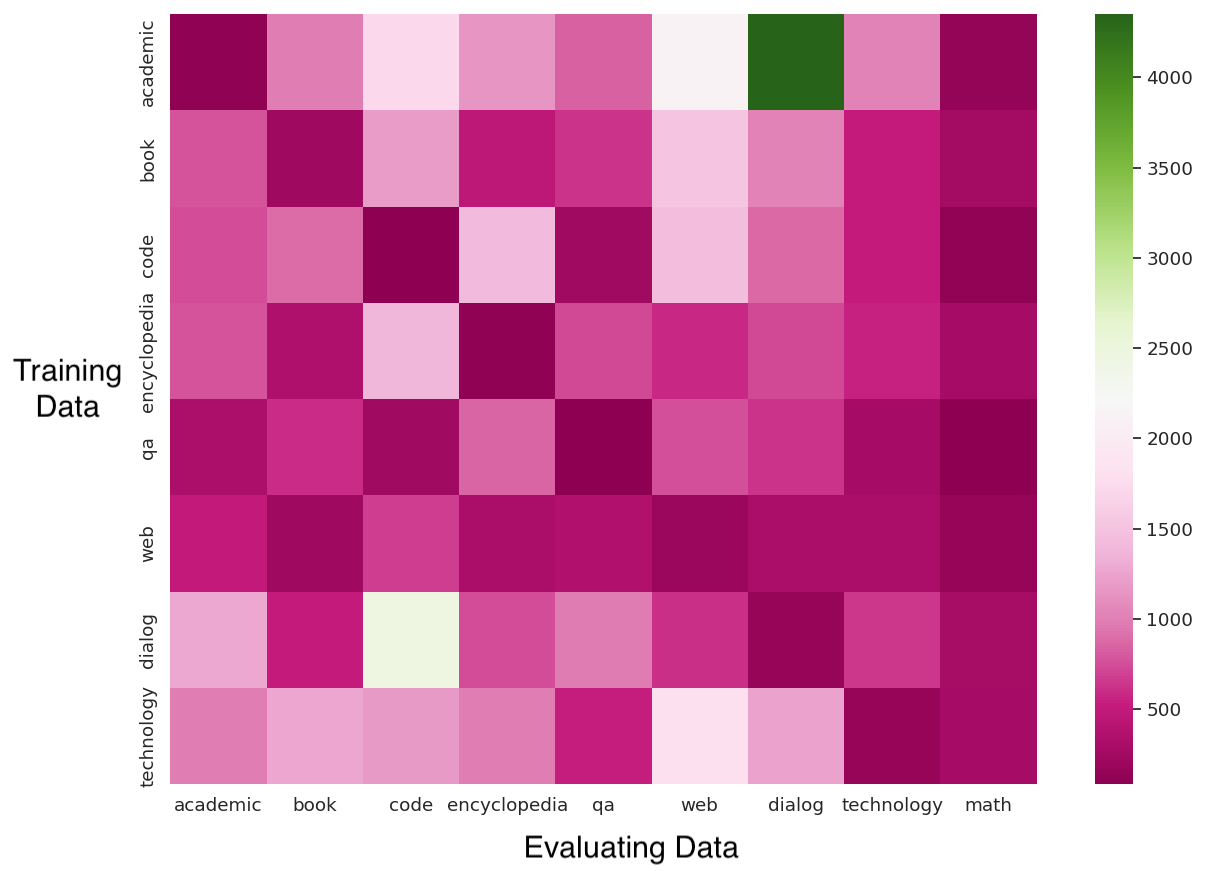}
    \caption{The perplexity of model train on single-source data and evaluate on other data. The higher perplexity indicates the model trained on source data cannot perform well on the evaluation data.}
    \label{fig:heatmap}
\end{figure}

Further evaluations explore the performance of a model trained on a single data source when extended to other sources, probing correlations between them. As depicted in Figure~\ref{fig:heatmap}, varying levels of generalization emerge from different datasets. Webtext, encyclopedia, and QA data demonstrate robust generalization across other sources, indicating their content encompasses diverse information across various domains. In contrast, models trained on academic and code data exhibit adept performance on math data but offer weaker generalization, possibly due to domain specificity and unique formatting information. Based on these analyses, the final dataset mixture ratio incorporates several principles:

\begin{enumerate}
    \item Maintain a high proportion of quality web text and encyclopedia data given their diversity.
    \item Downsample the ratio of math data to prevent overfitting.
    \item Enhance mathematical capabilities using code and academic data, while carefully mitigating the influence of format through diversified sampling.
    \item Retain some dialogue and book data, useful for learning long-range dependencies.
\end{enumerate}

In addition to the mix ratio, the data curriculum or training order are vital factors in model training. Previous research employs diverse curricula for specific skills. For instance, in the context of models designed for code capability, StarCoder~\citep{li2023starcoder} and AlphaCode~\citep{li2022competition} exclusively train models on code data, while CodeX~\citep{chen2021evaluating} and CodeLLaMA~\citep{roziere2023code} train models on code data after training on text data. \citet{chen2023skill} demonstrates that different sources of data induce distinct skills, and training with a specific data ordering can result in faster learning compared to training on skill-specific data. 

In contrast to~\citet{chen2023skill}, this research does not concentrate on data curricula but seeks to identify periodic patterns in training loss resulting from non-uniform data shuffling. An experiment is conducted, comparing the performance of models trained using different schedules. Two datasets are processed, each containing identical data but following distinct curricula. One utilizes uniform shuffling, and the other is ordered according to the data source. Experimental outcomes indicate that early knowledge acquisition is comparably consistent for both non-uniform and uniform data, though non-uniform data exhibits signs of forgetfulness latter. However, when samples from the same data source are grouped, performance improves in in-context learning~\citep{brown2020language} scenarios. It is hypothesized that when samples are drawn from the same data source, specific data distributions enhancing the model's contextual focus are likely to emerge. More detailed insight into the experiment can be found in Section~\ref{sec:stable}.

\section{Our Model}
In this section, we propose a series of lightweight language models MindLLM, a custom design that amalgamates various beneficial technologies and modules to ensure stable training and optimal performance.

\subsection{MindLLM-3B}
\label{subsec:Blue Space}
MindLLM-3B employs a decoder-only Transformer architecture, implementing several significant improvements.

\paragraph{Embedding} 
Our first step involves introducing a normalization step to stabilize the input embeddings. Following this, we replace the learned absolute positional embeddings with rotary position embeddings (RoPE)~\citep{su2021roformer} at each network layer. RoPE represents a static form of relative positional embeddings, modifying the embedding space to linearly depend on the attention of a token at position $m$ to a token at position $n$ on the difference $m-n$. This results in valuable features such as flexibility in handling varying sequence lengths, natural decay in inter-token dependency with increased relative distance, and enhanced linear self-attention with relative position encoding. These embeddings are applied to every embedding vector following the RoFormer approach.




\paragraph{Normalization}
Instead of the traditional LayerNorm operator~\citep{ba2016layer}, we utilize Root Mean Square Layer Normalization (RMS Norm)~\citep{zhang2019root}. This has been shown to provide performance comparable to LayerNorm but with significantly reduced running time, an essential characteristic for efficiently training large language models. In addition, we implement DeepNorm~\citep{wang2022deepnet} to adjust the residual connection in the Transformer layer with theoretically derived initialization. This combines the advantages of Post-Norm and stable training of Pre-Norm, facilitating the training of deeper models.




\paragraph{Attention}
In our quest to expedite training, we implement an efficient version of the causal multi-head attention operator, drawing inspiration from various sources such as~\citet{dao2023flashattention,dao2022flashattention}. Our choice, FlashAttention-2~\citep{dao2023flashattention}, provides optimal work partitioning to reduce memory usage and computation. This optimization involves algorithmic adjustments to minimize non-matmul FLOPs, parallelizing attention computations across different thread blocks, and appropriate work distribution among warps.

\paragraph{Feed-Forward Layer}
To further enhance performance, we substitute the GeLU non-linearity with the GeGLU activation function~\citep{shazeer2020glu}. Variants of Gated Linear Units in the feed-forward sublayers of Transformers have demonstrated improvements compared to standard activations. We maintain a weight matrix dimension of 8/3d to ensure consistency of total parameters with the standard feed-forward layer.




MindLLM-3B contains 3.1 billion parameters based on the proposed architecture, trained from scratch, on over 500 billion tokens. The parameter details are shown in Table~\ref{tab:parameters}. We use byt-pair encoding (BPE)~\citep{shibata1999byte} from SentencePiece~\citep{kudo2018sentencepiece} to tokenize the data and the final vocab size of our tokenizer is 125,700.

\begin{table}[tb]
    \centering
    \resizebox{\textwidth}{!}{
    \begin{tabular}{cccccc}
    \hline
         \textbf{Model} & \textbf{Hidden size} & \textbf{FFN Size} & \textbf{Heads Num.} & \textbf{Layers Num.} & \textbf{Sequence Length}  \\ \hline
         MindLLM-1.3B & 2048 & 8192 & 16 & 24 & 1024 \\
         MindLLM-3B & 2560 & 8192 & 32 & 32 & 2048 \\ \hline
    \end{tabular}
    }
    \caption{Model details of MindLLM.}
    \label{tab:parameters}
\end{table}

\subsection{MindLLM-1.3B}
\label{susec:B-GPT-NEO}
MindLLM-1.3B adopts the architecture of GPT-Neo-1.3B which employs a decoder-only Transformers architecture~\citep{vaswani2017attention}, adhering to the design principles of the OpenAI GPT model~\citep{radford2018improving}with minor modifications. These include the local-attention~\citep{child2019generating}pre-normalization, final normalization, and GeLU non-linear activation function~\citep{hendrycks2016bridging}.

Existing open-source LLMs (e.g. Llama, Alpaca) recognize Chinese scripts weakly, so we extend the LLMs tokenizer and LLMs word embeddings matrix with additional Chinese tokens. We train a SentencePiece ~\citep{DBLP:conf/emnlp/KudoR18} Chinese tokenizer using the Wudao corpora and then integrate the vocabulary from the Chinese tokenizer into the original LLM tokenizer. Furthermore, we augment the word embeddings by resizing the original word embeddings matrix, which has a shape of $ V \times D $, to a new shape of $ V^{\prime} \times D $ by appending new rows. Here, $D$ represents the hidden size, $V$ is the original vocabulary size, and $V^{\prime}$ is the new vocabulary size.

\section{Pre-training}
\label{sec:pretrain}
During the pre-training stage, we employ two distinct pre-training strategies: (1)training on bilingual mixture data and (2)training on monolingual data then transfer(Section~\ref{pretrain_detials}). The two distinct pre-training strategies underscore the influence of different data curricula on the acquisition of multilingual capabilities. To evaluate the pre-trained model's performance, we conduct testing and comparisons between MindLLM and other models of equivalent or larger scale (Blooms~\citep{Bloom}, LLaMA~\citep{touvron2023llama}, MOSS~\citep{sun2023moss} and  Baichuan~\citep{baichuan2023baichuan2}) using standard benchmarks (Section~\ref{Over_all_evlauation}). Concurrently, we conduct in-depth evaluations of the pre-trained models' capabilities based on open-source frameworks and datasets (Section~\ref{capability_evaluation}). Furthermore, we analyze the impact of different data construction approaches and pre-training frameworks on models' performance (Section~\ref{sec:stable}). These outcomes underscore the potential of lightweight LLMs and offer valuable insights and crucial guidance for the pre-training of such models.
\subsection{Training Objective and Technology}

MindLLM is trained using Hugging Face transformers frameworks\citep{wolf-etal-2020-transformers}, which offer a wide range of tools and modules for effectively customizing models and the pre-training process. 
Additionally, DeepSpeed~\citep{10.1145/3394486.3406703} is employed to facilitate the training process. It provides ZeRO~\citep{10.5555/3433701.3433727} optimizer and parallelism technologies. For the tradeoff between memory consumption and training speed, we use data parallelism and ZeRO stage 1, with which only optimizer states are partitioned across the processes. 
We adopt the bfloat16 format to keep the training with stability.

\subsection{Pre-training Strategies}
\label{pretrain_detials}
We pre-train our MindLLM models from scratch in two different learning strategies for bilingual language capability. The comparison in different pre-training strategies reveals the impact of  data curriculum for language capacity.

\paragraph{Train on Bilingual Mixture Data}
For our first strategy, we train MindLLM-3B directly on a bilingual mixed dataset. The appropriate proportion mixture of Chinese and English data is roughly 1:1.5, as detailed in Table~\ref{tab:train_data_all_model}.For optimizer, we use AdamW~\citep{AdamW} with $\beta_1=0.9,\beta_2=0.95$ and $\epsilon=1e^{-8}$ and initiate a 16000-step warm-up process for MindLLM to reach the max learning rate, followed by \textit{linear} decay to decrease the learning rate to 10\% of the peak value during the total 80000 training steps. We use a weight decay of
0.1 and clip the grad norm to 1.0.
\begin{table}[h]
    \centering
    \resizebox{\linewidth}{!}{
    \begin{tabular}{ccccccc}
    \hline
         \textbf{Model}&\textbf{Stage}&\textbf{Scheduler}&\textbf{Max LR}&\textbf{Min LR}&\textbf{\makecell[c]{Batch\\Size} }&\textbf{Tokens}\\
    \hline
         \multirow{2}{*}{\textbf{MindLLM-1.3B}}&Monolingual Training&Constant&$5.0\times10^{-5}$& $5.0\times10^{-5}$&1536&159B \\
         \cdashline{2-7}
         &\makecell[c]{Transfer Training}&\makecell[c]{4k-step warmup\\linear decacy}&$5.0\times10^{-4}$&$5.0\times10^{-5}$&1512&164B\\
         \cdashline{2-7}
    \textbf{MindLLM-3B}&Bilingual Training&\makecell[c]{16k-step warmup\\linear decacy}&$3.0\times10^{-4}$&$3.0\times10^{-5}$&3072&500B\\
    \hline
    \end{tabular}
    }
    \caption{Training details of MindLLMs.}
    \label{tab:training_details}
\end{table}

\paragraph{Train on Monolingual Data then Transfer}
For the second training strategy, we train our model on English data initially and then expand the vocabulary and train on Chinese data for language transferring. This strategy is common used in language transferring for large language model~\citep{cui2023efficient}. Specifically, we initially train the MindLLM-1.3B using Pile dataset~\citep{gao2020pile}. Subsequently, we expand the model's vocabulary by incorporating Chinese vocabulary and train the model on a bilingual language dataset with approximately a 1:1 ratio of Chinese and English data. This approach ensures that the model retains its previously learned English knowledge while acquiring Chinese language capabilities. We use AdamW~\citep{AdamW} in botch stages with $\beta_1=0.9,\beta_2=0.999$ and $\epsilon=1e^{-8}$. Specifically, we conduct training over 101,100 steps in the first stage with a constant learning rate of $5.0\times10^{-5}$. When we involve Chinese data during the transfer training stage (with a total of 105,900 training steps), we extend the warm-up phase and increase the maximum learning rate to $5.0\times10^{-4}$. This adjustment is made to help the model acquire bilingual capacity while effectively align the Chinese and English knowledge to achieve an optimal balance. For specific details, please refer to Table~\ref{tab:training_details}

\subsection{Overall Evaluation Settings}
Here we present the evaluation method, while the specific details of baselines and benchmarks are introduced in the corresponding section.

We evaluate the pre-trained models under a few-shot setting as the in-context learning~\citep{brown2020language} is crucial for pre-trained models. Conversely, the models after instruction tuning are evaluated in a zero-shot setting as the models are expected to understand the instruction~\citep{Becoming_self_instruct}.
We evaluate our models on multiple-choice tasks and free-form generation tasks~\citep{DBLP:journals/corr/abs-2309-10305, Bloom}.
\begin{enumerate}
    \item \textbf{Free-form generation}: Models are provided with inputs and generate the results subsequently as used in tasks such as summarization and translation.
    \item \textbf{Multiple-choice}:  Models are provided with a question
and multiple choices and asked to select the
most appropriate candidates.
\end{enumerate}

Specifically, for multiple-choice questions, we employ a discriminative approach to extract the model's answer. For each candidate answer, we first calculate the perplexity (PPL) of the model individually and select the candidate with the lowest perplexity as the final choice. 

To ensure a fair evaluation, we primarily rely on two open-source evaluation frameworks, lm-evaluation-harness~\citep{lm_harness} and OpenCompass~\citep{2023opencompass}. Specifically, for the MMLU~\citep{MMLU} and C-Eval~\citep{ceval} tasks, we adhere to the official evaluation methods.

\subsection{Bilingual Alignment Optimize the Monolingual Performance}
\label{B2M}
We train MindLLM-1.3B (Section~\ref{susec:B-GPT-NEO}) in a two-stages training strategies, injecting a mixture of Chinese and English data when the model has developed a well-trained capability. Therefore, a question has arisen: \textbf{Does the mixture of Chinese data affect the model's English ability?} Our evaluation results show that the inclusion of Chinese data can improve English performance in the second training stage with only limited English data.

We compare our model with GPT-Neo-1.3B which shares the identical structure and training data source~\citep{gao2020pile}. As shown in able~\ref{tab:training_data_neo}, the main difference is the training data where the GPT-Neo-1.3B is trained on 380 billion English-only tokens but MindLLM-1.3B is trained on 241 billion English tokens and 82 billion Chinese tokens.
\begin{table}[h]
    \centering
    \begin{tabular}{cccccc}
         \hline
         \multirow{2}{*}{\textbf{Model}}&\multicolumn{3}{c}{\textbf{Pre-training Data}}\\
         \cline{2-4}
         &\textbf{En}&\textbf{Zh}&\textbf{Total}\\
         \hline
         GPT-Neo-1.3B&380B &- &380B\\
         MindLLM-1.3B&241B&82B&323B\\
         \hline
    \end{tabular}
    \caption{Pre-trianing data comparison of GPT-Neo and MindLLM-1.3B.}
    \label{tab:training_data_neo}
\end{table}

\paragraph{Evaluation Setting}We use \textbf{MMLU}~\citep{MMLU}, the Massive Multitask Language Understanding which which comprises a diverse set of multiple-choice questions spanning various academic subjects to evaluate the two models' common skills of English. We employed the official evaluation methodology in 5-shot setting.

\paragraph{Results and Analysis}

The results are presented in Table~\ref{tab:mmlu_result_neos}. In comparison to GPT-Neo,  MindLLM-1.3B exhibited superior average performance (26.6 vs 24.1) in English tasks with much smaller training data. Particularly impressive is its performance on the Social Sciences task, where MindLLM-1.3B achieved a score of 29.2 compared to 22.3 for GPT-Neo.

\begin{table}[h]
    \centering
    \begin{tabular}{cccccc}
    \hline
         \textbf{Model}&\textbf{Humanities}&\makecell[c]{\textbf{Social} \textbf{Sciences}}&\textbf{STEM}&\textbf{Other}&\textbf{Average} \\
    \hline
         GPT-Neo-1.3B &24.8	&22.3	&25.3	&23.9	&24.1\\
         MindLLM-1.3B&\textbf{24.9} &\textbf{29.2}	&\textbf{27.6} &\textbf{25.8} &\textbf{26.6}\\
    \hline
    \end{tabular}
    \caption{Five-shot results of GPT-neo-1.3B and our MindLLM-1.3B on MMLU. }
    \label{tab:mmlu_result_neos}
\end{table}

The total amount of bilingual data (323B) in MindLLM-1.3B is less than the amount of English data (380B) used in GPT-Neo. Hence, we posit that the incorporation of Chinese data enhances the model's performance of English tasks. This improvement could be attributed to the information shared between different languages during the second training stage.
\subsection{Dose Unstable Loss Impact the Performance?}
\label{sec:stable}
In this section, our objective is to examine the impact of loss instability on model performance during training, as it relates to different data construction methods. We begin with a concise overview of two distinct data construction approaches, \textbf{Blocked} and \textbf{Shuffled}. Subsequently, we analyze the advantages and shortcomings of the two kinds of data, and give some suggestions for data construction.
\subsubsection{Data Construction Methods: Blocked v.s. Shuffled}
In this experiment, we maintain the consistency of the remaining parameters, such as the data size and training hyperparameters, while employing two different training data construction methods for model pre-training:
\begin{itemize}    
    \item \textbf{Blocked}: In this setting, we classify the bilingual data data separately based on meta-categories, and subsequently arrange them in a specific category order. During the model training process, Chinese and English data are sequentially presented to the model in an interleaved manner, ensuring balanced training across different languages.
    \item \textbf{Shuffled}: In this setting, we uniformly shuffle all categories of data, aiming to ensure that the model sees a uniform distribution of Chinese and English data in each training batch.
\end{itemize}
\begin{figure}[h]
    \centering
    \includegraphics[scale=0.3]{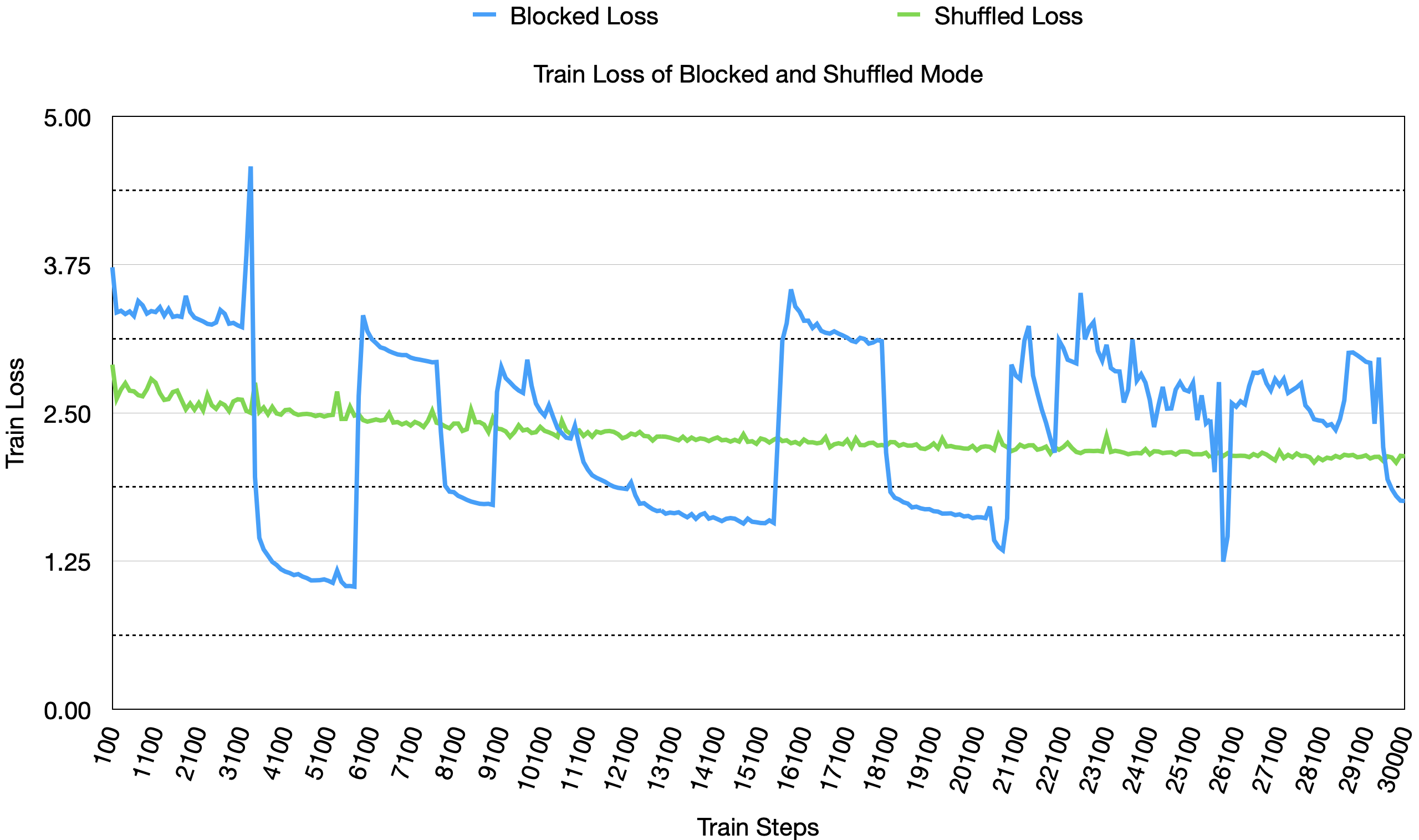}
    \caption{Training loss of different settings. \textcolor{green}{Green} means \textbf{Shuffled}, which has a stable training process. \textcolor{blue}{Blue} means \textbf{Blocked}, whose training process is unstable.}
    \label{fig:trian_process_stability}
\end{figure}
\paragraph{Stable and Unstable Training Process}
As shown in the Figure~\ref{fig:trian_process_stability}, for the \textbf{Blocked} mode, different categories of data are fed to the model sequentially. The model learns in a category-by-category manner, like a curriculum learning process~\citep{Curriculum_Learning}. Consequently, the loss suddenly rises when the model encounters new category data, followed by a gradual decrease as it adapts to the new category. This dynamic causes the overall training process of the model to exhibit instability. For the \textbf{Shuffled} mode, the model's training loss exhibits a consistent and gradual decrease, indicating a stable training trend.

\subsubsection{Results and Analysis}
\begin{figure}[h]
    \centering
    \includegraphics[scale=0.3]{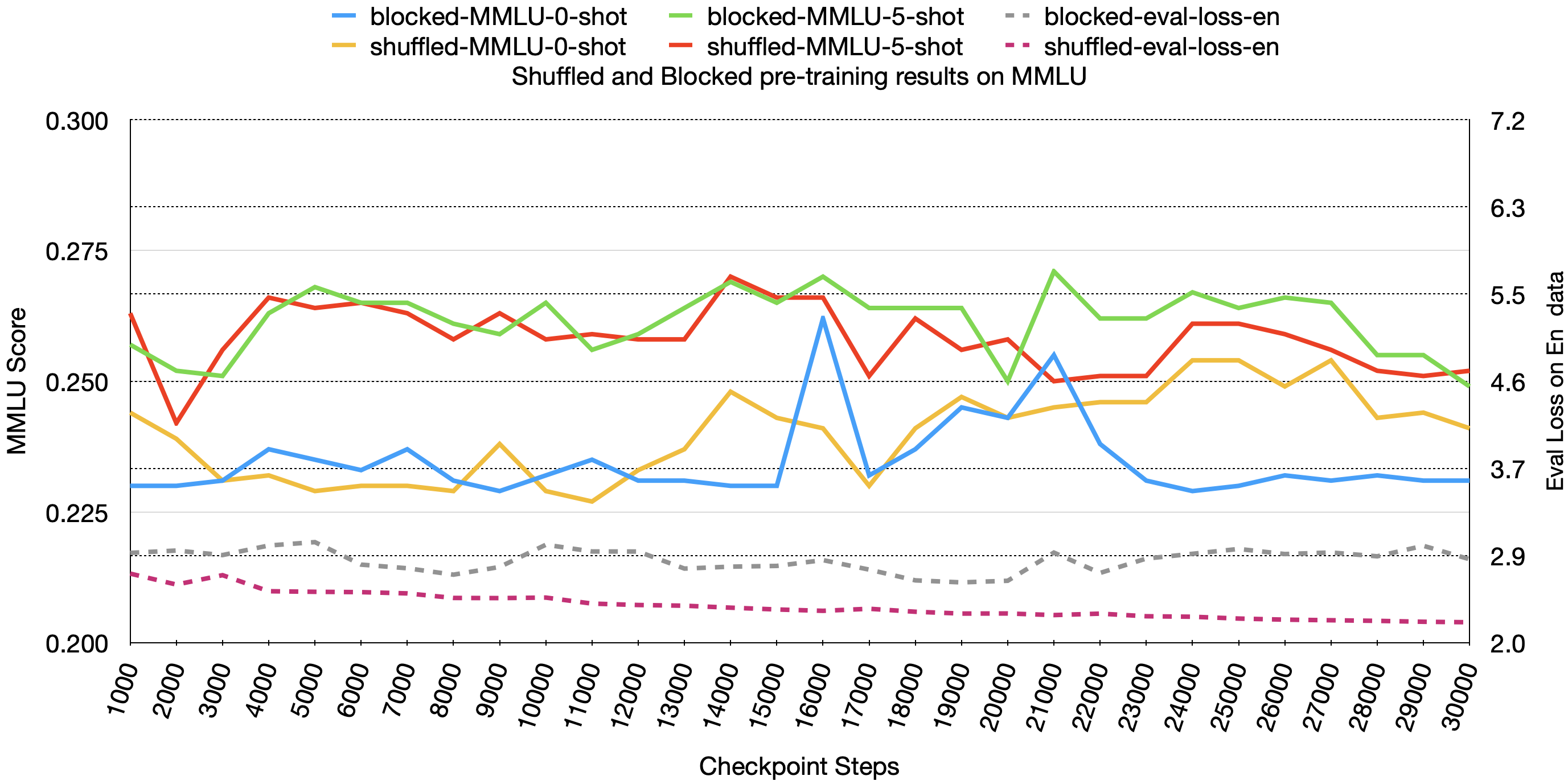}
    \caption{5\&0-shot performance on MMLU with validation loss on English data.}
    \label{fig:stability_mmlu}
\end{figure}

\begin{figure}[h]
    \centering
    \includegraphics[scale=0.3]{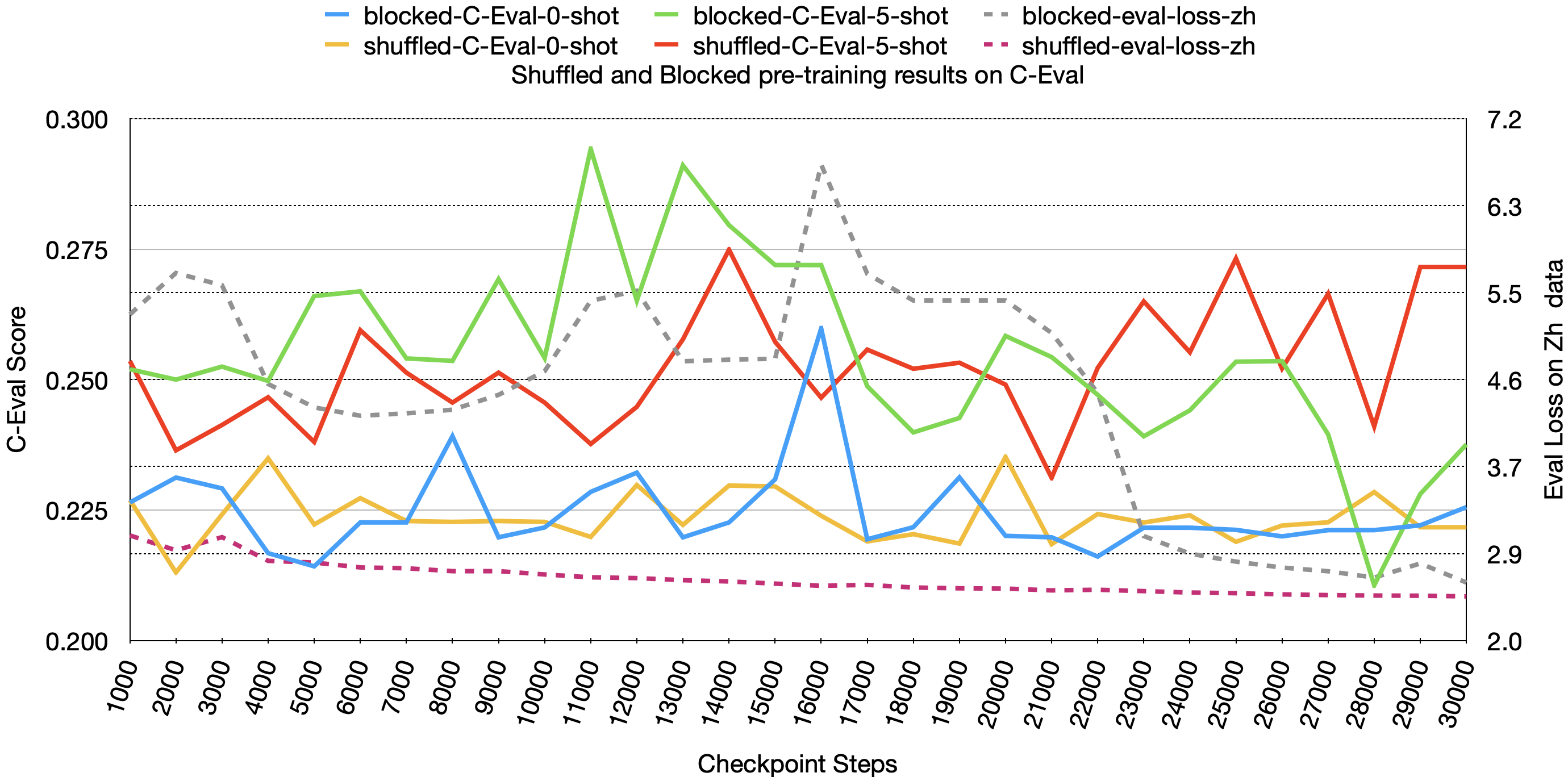}
    \caption{5\&0-shot performance on C-Eval with validation loss on Chinese data.}
    \label{fig:stability_ceval}
\end{figure}

During the training process, we utilize \textbf{C-Eval}~\citep{ceval}, a comprehensive Chinese evaluation benchmark consisting of over 10,000 multiple-choice questions and \textbf{MMLU} benchmark to evaluate the model's general abilities in Chinese and English, respectively. Specifically, we utilize C-Eval's validation set for efficiency evaluation. Additionally, we compute the validation loss of the model on both Chinese and English data throughout the entire training process. The validation set is obtained by uniform sampling data from all data categories. Results are shown in Figure~\ref{fig:stability_mmlu} and Figure~\ref{fig:stability_ceval}.


\paragraph{Comparable Problem Solving Performance in Stable and Unstable Training}
\label{para:Comparable_Problem_Solving_Memorization}
The zero-shot result provides some insights into the model's capacity to leverage pre-training knowledge to a certain extent. Based on the zero-shot results from MMLU and C-Eval benchmarks, the performance of the stably trained model (\textbf{Shuffled}) is on par with that of the unstably trained model (\textbf{Blocked}). This suggests that both models learn or memorize comparable English and Chinese knowledge during the training process.

\paragraph{Unstable Training Results in Forgetting}
Firstly, when considering the validation loss in both languages, the model trained with the \textbf{Shuffled} mode exhibits a consistent and gradual decrease in loss, maintaining a low value throughout (dashed \textcolor{purple}{purple} line in Figure~\ref{fig:stability_mmlu} and Figure~\ref{fig:stability_ceval}). On the contrary, the model trained with the \textbf{Blocked} mode displays higher and more variable validation loss (dashed \textcolor{gray}{gray} line in Figure~\ref{fig:stability_mmlu} and Figure~\ref{fig:stability_ceval}). These observations indicate that models trained with the shuffled mode are more adept at fitting the pre-trained data and have a higher likelihood of acquiring comprehensive underlying knowledge, which means not only the problem-solving knowledge mentioned above. What's more, during the training process, the model trained with the \textbf{Shuffled} mode exhibits a consistent improvement in 5-shot performance on C-Eval (solid \textcolor{red}{red} line in Figure~\ref{fig:stability_ceval}) and zero-shot performance on MMLU (solid \textcolor{orange}{orange} line in Figure~\ref{fig:stability_mmlu}). Inversely, the performance of the model trained with the \textbf{Blocked} mode exhibits notable inconsistency. In terms of zero-shot performance on MMLU, the \textbf{Blocked} trained model performs similarly to the model trained in the shuffled mode for the first 22,000 steps. However, beyond this point, its performance begins to decline while the Shuffled mode model continues to improve (solid \textcolor{blue}{blue} line in Figure~\ref{fig:stability_mmlu}). Similarly, in the 5-shot setting on C-Eval, the \textbf{Blocked} trained model initially outperforms the model trained in the shuffled mode up until approximately 17,000 steps, after which it experiences a sharp decline (solid \textcolor{green}{green} line in Figure~\ref{fig:stability_ceval}). We posit that this issue may be attributed to the 
\textbf{Blocked} approach employed for model training, resulting in the challenge of forgetting knowledge acquired in earlier phases.

\paragraph{What Factors Affect In-Context Learning in Unstable Training?}


Furthermore, during the blocked training process, we observe a consistently higher performance in the 5-shot setting on C-Eval in the early stages, while there is a sudden decline in the later stages. Prior works have posited the potential influence of particular data distributions on acquisition of the in-context learning~\citep{brown2020language} ability of a model~\citep{ICL_DATA_ditribution,ICL_concept}.  For instance, data exhibits entity clustering or the presence of entities with ambiguous attributes necessitating the model's adept utilization of contextual information~\citep{ICL_DATA_ditribution}.  Whether there is a potential relation between this phenomenon and the data distribution or data quality will be a focus of future investigations.

The aforementioned results highlight that the knowledge acquired by the model through stable training is likely to be more comprehensive, leading to overall performance stability. Although unstable training can result in model forgetting in later-stage, whether it contributes to the model's capacity to enhance in-context learning requires further investigation. These findings emphasize the importance of randomly shuffling the data to achieve a uniform distribution during the model's pre-training stage.

\subsection{Lightweight Models Beat Larger Ones}
\label{Over_all_evlauation}
In this section, we report the evaluation results of our models and some other models on standard benchmarks.
\subsubsection{Evaluation Settings}
\paragraph{Models}
Models that we compair with are as follows: MOSS-Base-16B~\citep{sun2023moss}, LLaMA-2-7B~\citep{DBLP:journals/corr/abs-2307-09288}, Baichuan2-7B~\citep{DBLP:journals/corr/abs-2309-10305}, LLaMA-7B~\citep{touvron2023llama}, MPT-7B~\citep{MosaicML2023Introducing}, Falcon-7B~\citep{refinedweb},  Bloom-3B\&7B~\citep{Bloom}, Open-LLaMA-3B\&7B~\citep{openlm2023openllama}, GPT-Neo-1.3B\&2.7B~\citep{gpt-neo} and GPT-J-6B~\citep{gpt-j}. The amount of data for training these models is shown in Table~\ref{tab:train_data_all_model}:
\begin{table}[h]
    \centering
    \begin{tabular}{lcrrr}
         \hline
         \multirow{2}{*}{\textbf{Model}}&\multirow{2}{*}{\textbf{Language}}&\multicolumn{3}{c}{\textbf{Pre-training Data}}\\ \cline{3-5}
         &&\textbf{En}&\textbf{Zh}&\textbf{Total}\\
         \hline 
         Baichuan2-7B&multilingual& &  &2600B\\
         LLaMA-2-7B&multilingual&1794B &2.6B  &2000B\\
         Falcon-7B&multilingual&- &-  &1500B\\
         
         LLaMA-7B&multilingual&- & - &1000B\\
         Open-LLaMA-3B\&7B&multilingual&- &- &1000B \\
         Bloom-7B\&3B&multilingual&110B &59.5B  &366B\\
         \hline
         MOSS-Base-16B&bilingual&375B &100B  & 700B\\
        \textbf{MindLLM-3B}&bilingual&285.7B &214.3B  &500B\\
         \textbf{MindLLM-1.3B}&bilingual&241B&82B&323B\\
         \hline
         MPT-7B&monolingual&1000B &-  &1000B\\
         GPT-J-6B&monolingual&402B &-  &402B\\
         GPT-Neo-1.3B\&2.7B&monolingual&380B &- &380B\\
         \hline
    \end{tabular}
    \caption{Pre-trian data comparison of different models}
    \label{tab:train_data_all_model}
\end{table}
\paragraph{Benchmarks}
We mainly use \textbf{MMLU}~\citep{MMLU} and \textbf{AGIEval}~\citep{AGIEVAL} to evaluate the model's general ability in English, \textbf{C-Eval}~\citep{ceval} and \textbf{CMMLU}~\citep{CMMLU} to evaluate the model's Chinese ability. Specifically, for \textbf{AGIEval}, we only use the English part for testing in 4-shot setting derived from OpenCompass~\citep{2023opencompass}. For C-Eval and MMLU, we adopt the official implementations and 5-shot for evaluation. For CMMLU,  we adopt 4-shot evaluation derived from lm-evaluation-harness~\citep{lm_harness}.

\subsubsection{Overall Performance on Standard Benchmarks}
Few-shot results on standard benchmarks are shown in Table~\ref{tab:overall_result_pt}. Our evaluation results demonstrate that MindLLMs can match or even surpass the performance of larger models in both Chinese and English.

\begin{table}[h]
    \centering
    \begin{tabular}{lccccc}
    \hline
    \textbf{Model}&\textbf{MMLU}&\textbf{AGIEval (En)}&\textbf{C-Eval}&\textbf{CMMLU}&\textbf{Average} \\
    \hline
         Baichuan2-7B& 54.2 &23.08 & 54.70 & 56.82& 47.20\\
        LLaMA-2-7B & 45.6&23.21  & 28.90& 32.88&32.65\\
         LLaMA-7B &34.9	&23.03 &26.70  &26.77 &27.85 \\
         Open-LLaMA-7B &29.9 &22.94 &25.90&25.69 &26.11\\
         \textbf{MindLLM-1.3B} &26.6 &23.02	&26.10 &25.33 &25.26\\
         Bloom-3B &26.3	&22.83	&25.90	&25.69	&25.18\\
         \textbf{MindLLM-3B} &26.2 &22.86	&25.70 &25.00 &25.02\\
         MOSS-Base-16B &22.3 &22.90	&26.60 &27.68 &24.87\\
         Open-LLaMA-3B &26.7 &22.73	&24.20 &25.60 &24.81\\
         Bloom-7B&25.5	&22.99	&22.80	&26.40	&24.42\\
         \hline
         \hline
         \textbf{Model}&\textbf{MMLU}&\textbf{AGIEval (En)}&-&-&\textbf{Average(En)} \\
    \hline
         Falcon-7B& 27.0&23.10 & -&- &25.05\\
         \textbf{MindLLM-1.3B} &26.6 &23.02	&- & -&24.81\\
         \textbf{MindLLM-3B} &26.2 &22.86&- &- &24.53\\
         GPT-J-6B &25.0	&23.11	&-	&-	&24.06\\
         MPT-7B& 26.4&21.44  & -&- &23.92\\
         GPT-Neo-2.7B &25.7 &22.10 &- &- &23.9\\
         GPT-Neo-1.3B &24.1	&22.07	&-	&-	&23.09\\
    \hline
    \end{tabular}
    \caption{Overall results of our models compared with other LLMs on standard benchmarks. The results at the bottom are for monolingual models, and the results at the top are for multilingual models.}
    \label{tab:overall_result_pt}
\end{table}
\paragraph{Performance on English Benchmarks} 
For MMLU, MindLLM-1.3B achieves an average score of 26.6, surpassing bigger models, such as GPT-Neo-2.7B (25.7), Bloom-3B (26.3), GPT-J-6B (25.0), Bloom-7B (25.5), MPT-7B (26.4)  and MOSS-Base-16B (22.3), and comparable with the performance of Open-LLaMA-3B (26.7). MindLLM-3B model (26.2) exhibits similar behavior to the 1.3B model.
For AGIEval (En), MindLLM-1.3B achieved an average score of 23.02, surpassing MOSS-Base-16B (22.90), MPT-7B (21.44), Bloom-7B (22.99), Open-LLaMA-7B (22.94), Bloom-3B (22.83), Open-LLaMA-3B (22.73),  and GPT-Neo-2.7B (22.10), approaching the performance of LLaMA-7B (23.03). As depicted in the bottom section of Table~\ref{tab:overall_result_pt}, MindLLMs' average English proficiency score surpasses that of the MPT-7B and GPT-J-6B, and is on par with that of the Falcon-7B. The latter models are characterized by larger scale and trained with a more extensive corpus of English data. 

\paragraph{Performance on Chinese Benchmarks}
In the 5-shot setting on C-Eval benchmark, MindLLM-1.3B and MindLLM-3B attain scores of 26.1 and 25.7, respectively, surpassing or equating to larger language models like  Open-LLaMA-7B (25.9), Bloom-7B (22.80), Bloom-3B (25.9) and Open-LLaMA-3B (24.2). As for CMMLU benchmark, MindLLM-1.3B and MindLLM-3B attain scores of 25.33 and 25.00, respectively, which is on par with the performance of Bloom-3B and Open-LLaMA-3B. What's more, although the MOSS-Base-16B exhibits subpar performance on MMLU and AGIEval (En), it demonstrates a higher performance in Chinese, achieving 26.6 and 29.6 on C-Eval and CMMLU, respectively. The above results indicate that MindLLMs has achieved a foundational level of proficiency in Chinese, while there is room for improvement in our model's performance. 

It is noteworthy that MindLLMs are still under training using better quality Chinese and English datasets based on our evaluation results. We will release the updated results in the future.

\subsection{Systematic Capability Analyses in Pre-training Evaluation}
\label{capability_evaluation}


In this section, we aim to evaluate the diverse capabilities of pre-trained models, such as \textit{Bilingual Capability, Complex Reasoning}. 
\begin{table}[h]
    \centering
    \begin{tabular}{ccc}
    \hline
    \textbf{Capability}&\textbf{Sub-Capability}&\textbf{Tasks}\\
    \hline
    \multirow{3}{*}{\textbf{Mathematics}}&  Arithmetic&Arithmetic~\citep{brown2020language}\\
    \cdashline{2-3}
    &General&\makecell{GSM8K~\citep{GSM8k}\\MATH~\citep{MMLU}}\\
    \hline
    \multirow{6}{*}{\textbf{\makecell{Complex\\Reasoning}}}&\makecell{Linguistic Reasoning}&\makecell{HellaSwag~\citep{HellaSwag}\\WinoGrande~\citep{WinoGrande}}\\
    \cdashline{2-3}
    &Logic Reasoning&LogiQA~\citep{LogiQA}\\
    \cdashline{2-3}
    &\makecell{Knowledge Reasoning}&\makecell{PubMedQA~\citep{PubMedQA}\\PIQA~\citep{PIQA}\\MathQA~\citep{MathQA}}\\
    \cdashline{2-3}
    &\makecell{Integrated Reasoning}&BBH~\citep{BIG-bench}\\
    \hline
    \textbf{\makecell{Bilingual\\Capability}}&-&Flores-101~\citep{flores}\\
    \hline
    \multirow{3}{*}{\textbf{Values}}&Truthfulness&TruthfulQA~\citep{TruthfulQA}\\
    \cdashline{2-3}
    &Toxicity&ToxiGen~\citep{hartvigsen-etal-2022-toxigen}\\
    \cdashline{2-3}
    &Ethics&Ethics~\citep{Ethics}\\
    \hline
    \end{tabular}
    \caption{Correspondence between Capability and Task of Capability Evaluation System.} 
    \label{tab:task_capability_mapping}
\end{table}
\paragraph{Capability Evaluation System}In this section, our primary focus lies in the evaluation and analysis of the model's English language proficiency. Since the optimization of MindLLM's Chinese data is currently underway, we will use SuperCLUE~\citep{SuperCLUE} benchmark to evaluate the Chinese capabilities of different models in the future. To evaluate the model's English capabilities, leveraging the OpenCompass~\citep{2023opencompass} and lm-evaluation-harness~\citep{lm_harness} open-source frameworks, we carefully select various evaluation tasks to assess different capabilities of these models. The capabilities and corresponding tasks are presented in Table~\ref{tab:task_capability_mapping}. Arithmetic ability assesses the model's proficiency in numerical computation, while the evaluation of general mathematical ability encompasses a broad spectrum of mathematical domains. Within complex reasoning ability, we subdivide it into linguistic reasoning, logical reasoning, knowledge reasoning, and integrated reasoning to gauge the model's reasoning competence in language, logic, and knowledge utilization. We also conduct evaluations for bilingual capability in Chinese and English to assess alignment in both languages for different bilingual or multilingual models. Lastly, we conduct security evaluations of the above models, which we report results in Appendix~\ref{sec:evaluation_values}. For each capability, we take the average performance of the corresponding tasks as the evaluation score. 



\subsubsection{Mathematics}
\paragraph{Evaluation Settings} Specifically, we employ the \textbf{MATH}~\citep{MMLU}, which contains 12,500 mathematical questions that are harder to solve, and \textbf{GSM8K}~\citep{GSM8k} datasets to evaluate the model's general mathematical performance, while the \textbf{Arithmetic}~\citep{brown2020language} datasets are utilized to assess its arithmetic capability. We use lm-evaluation-harness to implement evaluation on \textbf{Arithmetic} (5-shot) and OpenCompass to test models on \textbf{GSM8K} (4-shot) and \textbf{MATH} (4-shot) . Results on mathematics are shown in Table~\ref{tab:math_capability}.
\begin{table}[h]
    \centering
    \begin{tabular}{lcccc}
    \hline
        \textbf{Model} &\textbf{Arithmetic}&\textbf{GSM8K}&\textbf{MATH}&\textbf{Average} \\
    \hline
    Baichuan2-7B &87.12 &25.09 &5.32 &39.18\\ 
    LLaMA-2-7B  &76.33 &16.53 &3.28 &31.94\\   
    LLaMA-7B &68.85 &9.86 &2.96 &29.89\\
    Open-LLaMA-7B &53.60 &5.84 &2.30 &20.58\\
    Falcon-7B &45.76 &5.46 &1.62 &17.61\\
    \textbf{MindLLM-3B} &44.75 &1.21  &2.06 &16.01\\
    MOSS-Base-16B &37.82 &6.90* &2.40* &15.71\\
    MPT-7B &28.26 &9.10* &2.90* &13.42\\
    GPT-J-6B  &32.12 &5.31 &2.02 &13.15\\
    Open-LLaMA-3B &32.86 &2.05 &1.98 &12.30\\
      Bloom-7B  &2.84 &3.18 &0.50 &2.17\\
     GPT-Neo-2.7B   &2.86 &1.82 &1.38  &2.02\\
     \textbf{MindLLM-1.3B} &0.71 &2.65 &1.26 &1.54\\
     GPT-Neo-1.3B   &1.71 &1.36 &1.50&1.52\\
     Bloom-3B &1.16 &1.67 &0.64 &1.16\\
    \hline
    \end{tabular}
    \caption{Results on Arithmetic (5-shot), GSM8K (4-shot) and MATH (4-shot). * denotes
results derived from the leaderboard of OpenCompass~\citep{2023opencompass}}
    \label{tab:math_capability}
\end{table}
\paragraph{}
\paragraph{Great Potential of Lightweight LLMs in Mathematics}
As shown in Table~\ref{tab:math_capability}, we observe that models exhibiting a high average level of mathematical performance (more than 15 points) are predominantly the 7B models, with the exception of MindLLM-3B. Meanwhile, MindLLM-3B outperforms MPT-7B (13.42), GPT-J-6B (13.15) and MOSS-Base-16B (15.71) achieving an average score of 16.01, which demonstrates the significant potential of lightweight models to excel in mathematical performance. Specifically, MindLLM-3B achieves a score of 44.75 on the \textbf{Arithmetic} task, outperforming larger models like MOSS-Base-16B (37.82) and MPT-7B (28.26), and on par with the performance of Falcon-7B (45.76). MindLLM-3B further achieves an impressive score of 2.06 on the challenging \textbf{MATH} benchmark, surpassing Falcon-7B (1.62), GPT-J-6B (2.02), and Bloom-7B (0.5). As for MindLLM-1.3B, it achieves a score of 2.65 on \textbf{GSM8K}, a general mathematical ability evaluation benchmark, surpassing larger models like Bloom-3B (1.67), GPT-Neo-2.7B (1.82), and Open-LLaMA-3B (2.05).

\paragraph{Model Scale Limits Complex Capability Acquisition}
Nevertheless, MindLLM-3B's performance on GSM8K markedly deviates from that of other 7B models, as well as MindLLM-1.3B's performance on MATH and ARITHMETIC. Meanwhile, Open-LLaMA-3B exhibits similar behavior, yet it falls short of surpassing MindLLM-3B in terms of performance on ARITHMETIC or MATH. Hence, we posit that owing to the intricacies of  mathematics tasks, acquiring mathematical capability places substantial demands on a model's capacity. Typically, learning of broader mathematical knowledge necessitates a larger model capacity, while a smaller model might excel in a specific mathematical domain, albeit at the potential cost of performance in other aspects (such as the complex reasoning mentioned in Section~\ref{sec:complex_reasoning}, which is also a challenging task). As for the poor math average performance of Bloom-7B\&3B, it may be related to the consumption of model capacity on bilingualism mentioned in Section~\ref{Bilingual} and the pre-training data, in which math data was upsampled in MindLLM-3B's pre-training stage.

\subsubsection{Complex Reasoning}
\label{sec:complex_reasoning}
\paragraph{Evaluation Settings} Complex reasoning capability is one of the most important capabilities of a language model. We subdivide complex reasoning into linguistic reasoning (\textbf{HellaSwag}~\citep{HellaSwag}, \textbf{WinoGrande}~\citep{WinoGrande}), logical reasoning (\textbf{LogiQA}~\citep{LogiQA}), knowledge reasoning (\textbf{PubMedQA}~\citep{PubMedQA}, \textbf{PIQA}~\citep{PIQA}, \textbf{MathQA}~\citep{MathQA}) and integrated reasoning (\textbf{BBH}~\citep{BIG-bench}). All evaluations in this section are driven from lm-evaluation-harness in a 5-shot setting, except \textbf{BBH}, which we evaluate in a 3-shot setting.
\begin{table}[h]
    \centering
    \begin{tabular}{lccccc}
    \hline
         \textbf{Model}&\textbf{\makecell[c]{Linguistic \\Reasoning}}&\textbf{\makecell[c]{Logical\\Reasoning}}&\textbf{\makecell[c]{Knowledge\\Reasoning}}&\textbf{\makecell[c]{Integrated\\Reasoning}}&\textbf{Average}  \\
    \hline
    LLaMA-2-7B&66.05&29.34&59.77 &36.72&47.97\\
    Baichuan2-7B&62.08&29.19&60.62&37.73&47.41\\
    LLaMA-7B&64.38 &27.65 &60.03 &35.60&46.92\\
    Open-LLaMA-7B&60.57 &28.42 &54.95 &35.71&44.91\\
    Open-LLaMA-3B&57.06 &25.19 &54.81 &34.15&42.80\\
    MOSS-Base-16B&52.24&28.26&55.26&35.32&42.77\\
    Bloom-7B&55.94 &24.88 &53.09 &33.21&41.78\\
    \textbf{MindLLM-3B}&50.36 &23.20 &53.34 &34.45&40.33\\
    GPT-Neo-2.7B&51.38 &24.58 &50.96 &33.77&40.17\\
    Bloom-3B&49.43 &22.89 &50.35 &33.90&39.14\\
    GPT-Neo-1.3B&47.86 &24.12 &50.42 &33.40&38.95\\
    \textbf{MindLLM-1.3B}&40.03 &24.21 &46.49 &31.70&35.61\\
    \hline
    \end{tabular}
    \caption{Results on complex reasoning. 5-shot results for Linguistic, Logical and Knowledge Reasoning, and 3-shot results for Integrated Reasoning.}
    \label{tab:Complex_Reasoning}
\end{table}
\paragraph{Compromise between Multilingual Benefits and Model Capability Consumption}
Results are shown in Table~\ref{tab:Complex_Reasoning}. MindLLM-1.3B exhibits slightly inferior reasoning capabilities in comparison to GPT-Neo-1.3B (35.61 vs 38.95). This disparity is likely attributed to the pre-training of MindLLM-1.3B on both Chinese and English corpora. On one hand, this training approach yields monolingual benefits, as exemplified in Section~\ref{B2M}. However, it is important to note that, due to the model's limited scale, the acquisition of Chinese language proficiency consumes a portion of the model's capacity. As a result, for complex reasoning tasks, which are inherently difficult, the bilingual gains are outweighed by the associated model capacity costs. Consequently, while bilingual learning does confer certain competency improvements, it is imperative to strike a balance between capability improvement and capacity consumption. This assertion gains further support from the performance of Blooms and MindLLM-3B. Although their reasoning capability is not exceptional, their proficiency in bilingualism, as demonstrated in Section~\ref{Bilingual}, surpasses that of all other models. Furthermore, a larger pre-training dataset, potentially enriched with a broader range of world knowledge, can contribute to the augmentation of the model's reasoning capabilities. This is exemplified by Open-LLaMA-3B, which exhibits a superior scale of pre-training data compared to the pre-trained data of Bloom-3B and MindLLM-3B, as demonstrated in Table~\ref{tab:train_data_all_model}. As a result, smaller models still have the potential to reach the reasoning capability of larger models leveraged by the specific data.

\paragraph{How and When Does Code Data Affect the Reasoning Capabilities?}
\label{code_for_reasoning}
Previous works have demonstrated that code LLMs with larger model parameters are more effective than vanilla LLMs for reasoning~\citep{code1,code2,code3,code4}. MOSS-Base-16B~\citep{sun2023moss} is initialized with CodeGen~\citep{codegen} and further pre-trained on 100B Chinese tokens and 20B English tokens. Nevertheless, MOSS does not showcase notably strong reasoning skills, and we hypotheses that this could have contributed to the issue of forgetting~\citep{Continual_Learning_survey} or challenge of knowledge fusion~\citep{knowledge_fusion} during the second training stage. Hence, If code data indeed enhances the model's reasoning capabilities, future investigations should explore the optimal approach (when \& how) for its incorporation.
\subsubsection{Bilingual Capability}
\label{Bilingual}
We aspire for bilingual and multilingual models to achieve cross-linguistic comprehension and alignment, enabling these models to grasp shared linguistic characteristics and fostering the seamless assimilation of knowledge across diverse languages. Consequently, in this section, we predominantly investigate the model's capability for comprehension and alignment in both Chinese and English.
\paragraph{Evaluation Settings}
\textbf{Flores-101}~\citep{flores} is a translation benchmark used to evaluate model's multilingual capability. Here we select the En$\leftrightarrow$Zh part to compare the bilingual capability of different models in Chinese and English. We conduct this evaluation using OpenCampass in 8-shot setting. Specifically, we add Chinese-LLaMA-2-7B~\citep{Linly}, a variant of LLaMA-2-7B, designed for Chinese domain adaptation. This adaptation involves incorporating a substantial volume of mixed Chinese and English predictions, along with the inclusion of Chinese and English translation data in its second training stage. Detailed results are shown in Table~\ref{tab:flores}.
\begin{table}[h]
    \centering
    \begin{tabular}{lccccc}
    \hline
     \textbf{Model}&\textbf{Zh(S)-En}&\textbf{Zh(T)-En}&\textbf{En-Zh(S)}&\textbf{En-Zh(T)}&\textbf{Overall}\\
    \hline
    Bloom-7B&30.30&	29.65&	22.97&	0.11&	20.76\\
    Bloom-3B&	24.93&	27.46&	29.68&	0.16&	20.56\\
    LLaMA-7B&	39.14&	33.23&	1.01&	0.41&	18.00\\
    Open-LLaMA-7B&	26.00&	16.76&	0.93&	0.13&	10.96\\
    \textbf{MindLLM-3B}&	11.96&	10.22&	16.89&	0.48&9.89\\
    LLaMA-2-7B&	11.25&	14.83&	7.83&	0.79&	8.67\\
    MOSS-Base-16B&12.54&3.58&4.32&0.26&5.17\\
    Open-LLaMA-3B&	6.96&	8.22&	0.95&	0.17&	4.07\\
    Chinese-LLaMA-2-7B&2.90&7.86&0.92&0.09&2.95\\
    Baichuan2-7B&2.32&2.38&0.93&0.23&1.46\\
    \textbf{MindLLM-1.3B}&	0.71&	0.64&	0.95&	0.06&	0.46\\
    \hline
    \end{tabular}
    \caption{Detailed results of Flores-101 (En$\leftrightarrow$Zh) in 8-shot setting. \textbf{S} means Simplified Chinese, and \textbf{T} means Traditional Chinese.}
    \label{tab:flores}
\end{table}
\paragraph{Data Proportions Affect Bilingual Alignment}
Primarily, none of the aforementioned models exhibit strong performance in translating from English to traditional Chinese. This limitation can be attributed to the notably low proportion of traditional Chinese data in the pre-training datasets of these models. Furthermore, we observe that only Blooms and MindLLM-3B consistently excel in both Chinese-to-English and English-to-Chinese translation tasks, followed by LLaMA2-7B and MOSS-Base-16B. While LLaMA-7B and Open-LLaMA-7B demonstrate strong Chinese-to-English translation performance, they exhibit notable deficiencies in English-to-Chinese direction. Referring to Table~\ref{tab:train_data_all_model}, the ratio between English and Chinese pretrianing data in Blooms, and MindLLM3B is 1.85:1 and 1.3:1, respectively. In contrast, LLaMA-2-7B's pre-training data has a significantly imbalanced ratio of 687.5:1 between English and Chinese. The proportion of Chinese data in LLaMA-7B and Open-LLaMAs is even more limited. As a result, we have two hypotheses. First, LLMs can effectively learn a shared language representation based on an extensive volume of data in a particular language (\textit{source language}). Consequently, a smaller quantity of data in another language (\textit{target language}) becomes sufficient for understanding this language (\textit{target}) and achieving one-way alignment (target$\to$source), as exemplified by the performance of LLAMA-7B and Open-LLAMA-7B. Second, to enable the model to achieve a genuine understanding and alignment in both languages, a more equitable distribution of data from the two languages during training from scratch is advisable, as seen in Blooms and MindLLM-3B, which outperform all other models above in bilingual alignment capability, include LLaMAs, Open-LLaMAs, MOSS-Base-16B and Baichuan2-7B.
\paragraph{Injecting Capability through Language Transfer Training is Suboptimal}
Moreover, the performance of langauge transfer training for language alignment may be suboptimal, as indicated in the case of Chinese-LLaMA-2-7B in Table~\ref{tab:flores}. Even with the incorporation of a certain amount of Chinese and English translation data during Chinese domain adaptation~\citep{Linly}, the alignment performance from English to Chinese remains deficient. The above conclusion can also be generalized to the learning of multiple abilities, especially for such scale models with limited capacity. Furthermore, when considering the case of MOSS-Base-16B~\citep{sun2023moss}, which is initialized with  CodeGEN~\citep{codegen} as discussed in Section~\ref{code_for_reasoning} and dose not effectively integrate code to augment the it's reasoning capabilities. We can extrapolate this finding to encompass other capabilites acquired by the model. That is during the pre-training stage, we can equip the model with a range of capacities necessary for downstream tasks by judiciously allocating relevant data, thereby facilitating the early acquisition of target capabilities and knowledge fusion. Conversely, the method of augmenting a model's capacity through language transfer training is suboptimal, as the initial stages of learning have already consumed a significant portion of the model's finite capacity. Hence, during the ongoing training process, the incorporation of new capabilities and the fusion of existing and novel knowledge becomes a formidable challenge, even leading to forgetting~\citep{Continual_Learning_survey}. This could also explain the underperformance of Baichuan2-7B, LLaMAs, MOSS-Base-16B, and MindLLM-1.3B in terms of bilingualism. On one hand, the aforementioned models have exhibited better performance compared to Blooms and MindLLM-3B in some other capabilities, as demonstrated in Table~\ref{tab:overall_result_pt} and Table~\ref{tab:Complex_Reasoning}, thereby consuming the model's finite capacity. On the other hand, both MOSS-Base-16B and MindLLM-1.3B undergo langauge transfer training. Even with a more balanced distribution of Chinese and English data in the later stages, which can enhance the acquisition of bilingual proficiency, they still struggle to attain a high level of bilingual competence.

\subsection{Conclusion for Evaluation on Pre-training Stage}
In the pre-training stage, our evaluation and in-depth analysis of  different models on various general benchmarks and specific capabilities lead us to the following conclusions:
\begin{itemize}
    \item First, lightweight models exhibit great potential to match or even surpass the performance of larger models in specific capabilities and tasks. For instance, MindLLM-3B exhibits superior performance in mathematics even when compared to larger models with more extensive training data, such as MPT-7B, Falcon-7B and MOSS-Base-16B.  In the domain of bilingual capability, MindLLM-3B outperforms larger models like LLaMA2-7B and Baichuan2-7B.  This pattern also extends to general benchmarks, including MMLU and C-Eval.   
    \item Second, recognizing the model's limited capacity, we advocate the judicious inclusion of diverse data relevant to the downstream task's required capabilities during the early stages of pre-training. This approach facilitates the model's acquisition of the targeted capabilities and knowledge fusion. In contrast, the method of language or knowledge transfer training is suboptimal as existing knowledge occupies the model's finite capacity, hindering the assimilation of novel knowledge and the seamless integration of diverse information, even leading to catastrophic forgetting.
\end{itemize}

\section{Instruction Tuning}
\label{sec:instruction_tuning}
Instruction tuning can help LLMs better understand and response to requests given by humans, and has been proven to effectively improve the capabilities of LLMs\citep{DBLP:conf/nips/Ouyang0JAWMZASR22,DBLP:conf/iclr/WeiBZGYLDDL22,DBLP:conf/acl/WangKMLSKH23,DBLP:journals/corr/abs-2307-09288}. In this section, we aim to explore how to maximize the performance of lightweight models through instruction tuning. 

\subsection{Open-Source Instruction Tuning Data and Combination Strategy}
\label{sec:instruction_runing_combination}
Instruction tuning strategy is a key factor contributing to the remarkable success of LLMs. Nevertheless, there exists a scarcity of comprehensive experimental analysis focusing on instruction tuning data, particularly concerning lightweight models. Our objective is to undertake a deeper exploration into the impact of diverse quantities and types of instruction data on the performance of lightweight models. In this section, we provide a detailed description of the specific source and combination strategy of instruction tuning datasets used in each part of Section~\ref{sec:instruction_tuning}.

In Section~\ref{sec: multi_instruction_data}, we compile composite instruction tuning datasets in Chinese, English, and a blended bilingual dataset from various sources. The datasets are used for a comprehensive exploration into how the varying quantities and compositions of instruction tuning datasets influence the performance of lightweight models. The specific details of the datasets are provided in Table~\ref{tab: multi_datasets}.

\begin{table}[h]
\small
\centering
\begin{tabular}{cllr}
\hline
\bf Name                    & \bf Datasets               & \bf Sourced From                              & \bf Data Size \\ \hline
\multirow{7}{*}{\makecell{\bf MingLi\\(Chinese)}} & CLS\citep{DBLP:conf/coling/LiZ0S0MZ22}  & NLP datasets + Human-weitten & 396,031    \\
                         & COIG-Alignment\citep{DBLP:journals/corr/abs-2304-07987}         & NLP datasets + Generated LM               & 22,470     \\
                         & COIG-Exam\citep{DBLP:journals/corr/abs-2304-07987}              & Human-written from scratch                & 63,532     \\
                        & COIG-Leetcode\citep{DBLP:journals/corr/abs-2304-07987}          & Human-written from scratch                & 11,737     \\
                        & COIG-Translate\citep{DBLP:journals/corr/abs-2304-07987}         & NLP datasets + Translate                  & 66,858     \\
                         & Wizard-LM-Chinese~\citep{alpaca_luotuo}      & NLP datasets + Translate                  & 70,000     \\
                         & Chinese-Alpaca~\citep{chinese-llama-alpaca}         & NLP datasets + Translate                  & 52,002     \\ \hline
\multirow{11}{*}{\makecell{\bf Tulu\\(English)}}  & SuperNI~\citep{DBLP:conf/emnlp/WangMAKMNADASPK22}                & NLP datasets + Human-written & 96,913     \\
                         & CoT~\citep{DBLP:conf/nips/Wei0SBIXCLZ22}                    & NLP datasets + Human-written         & 100,000    \\
                         & Flan V2~\citep{DBLP:conf/icml/LongpreHVWCTZLZ23}                & NLP datasets + Human-written & 100,000    \\
                         & Dolly~\citep{DatabricksBlog2023DollyV2}                 & Human-written from scratch                & 15,011     \\
                         & Open Assistant 1~\citep{DBLP:journals/corr/abs-2304-07327}       & Human-written from scratch                & 34,546     \\
                         & Self-instruct~\citep{DBLP:conf/acl/WangKMLSKH23}          & Generated vanilla GPT3 LM              & 82,439     \\
                         & Unnatural Instructions~\citep{DBLP:conf/acl/HonovichSLS23} & Generated Davinci-002                  & 68,478     \\
                         & Alpaca~\citep{alpaca}                 & Generated Davinci-003                  & 52,002     \\
                         & Code-Alpaca~\citep{codealpaca}            & Generated Davinci-003                  & 20,022     \\
                         & GPT4-Alpaca~\citep{DBLP:journals/corr/abs-2304-03277}            & Generated Davinci-003 + GPT4           & 52,002     \\
                         & Baize~\citep{DBLP:journals/corr/abs-2304-01196}                  & Generated ChatGPT                      & 210,311    \\ \hline
\bf Bilingual                    & MOSS~\citep{DBLP:journals/corr/abs-2307-04964}                   & Generated Davinci-003                  & 1,074,551  \\ \hline
\end{tabular}
\caption{Chinese, English and Bilingual instruction tuning datasets investigated in this work.}\label{tab: multi_datasets}
\end{table}

\paragraph{Chinese Instruction Tuning Data - MingLi} For the Chinese dataset, We attempt to collect a representative sample of different styles of Chinese datasets, including datasets: (1) created by researchers from existing NLP datasets (CLS~\citep{DBLP:conf/coling/LiZ0S0MZ22}); (2) written by humans from scratch for the purpose of instruction tuning (COIG-Exam~\citep{DBLP:journals/corr/abs-2304-07987}, COIG-Leetcode~\citep{DBLP:journals/corr/abs-2304-07987}); (3) generated by proprietary models (COIG-Alignment~\citep{DBLP:journals/corr/abs-2304-07987}); (4) translated from existing NLP datasets (COIG-Translate~\citep{DBLP:journals/corr/abs-2304-07987}, Wizard-LM-Chinese~\citep{alpaca_luotuo}, Chinese-Alpaca~\citep{chinese-llama-alpaca}). After the selection process, we retain a total of 556,498 records as the Chinese instruction tuning dataset, named \textbf{MingLi}.

\paragraph{English Instruction Tuning Data - Tulu} For the English dataset, we utilize the open instruction-following dataset Tulu, constructed by \cite{DBLP:journals/corr/abs-2306-04751}. This dataset comprises various sub-datasets, including SuperNI~\citep{DBLP:conf/emnlp/WangMAKMNADASPK22}, CoT~\citep{DBLP:conf/nips/Wei0SBIXCLZ22}, Flan V2~\citep{DBLP:conf/icml/LongpreHVWCTZLZ23}, Dolly~\citep{DatabricksBlog2023DollyV2}, Open Assistant 1~\citep{DBLP:journals/corr/abs-2304-07327}, Self-instruct~\citep{DBLP:conf/acl/WangKMLSKH23}, Unnatural Instructions~\citep{DBLP:conf/acl/HonovichSLS23}, Alpaca~\citep{alpaca}, Code-Alpaca~\citep{codealpaca}, GPT4-Alpaca~\citep{DBLP:journals/corr/abs-2304-03277}, Baize~\citep{DBLP:journals/corr/abs-2304-01196}. To adapt to the lightweight model in this paper, we filter out multi-turn dialog data from the datasets and perform rule-based cleaning to remove some noisy data. Ultimately, we retain 417,421 records as the overall English training data.

\paragraph{Bilingual Instruction Tuning Data - MOSS}Additionally, due to the remarkable performance demonstrated by the MOSS~\citep{DBLP:journals/corr/abs-2307-04964} model, we also train our MindLLM-1.3B and MindLLM-3B lightweight models on their publicly available high-quality Chinese and English data corpus. The MOSS dataset encompasses seven categories: Brainstorming, Complex Instruction, Code, Role Playing, Writing, Harmless, and Others, comprising a total of 1,074,551 records.

\paragraph{Instruction Tuning on Tailored Data} In order to investigate whether a lightweight model can effectively enhance its specific capabilities through instruction tuning on specific datasets, we conduct experiments using two open-source available instruction tuning datasets. For the Chinese dataset, we utilize the "exam" portion of the WanJuan dataset~\citep{DBLP:journals/corr/abs-2308-10755}, which aims to help the model acquire foundational knowledge in subjects like textbooks and exam questions. This segment of data consists of 3,997,183 records. In the case of English, we employ the LogiCoT dataset~\citep{DBLP:journals/corr/abs-2305-12147} for instruction tuning, with the goal of enhancing the model's reasoning abilities. This segment of data comprises 668,488 records. Experimental details and specific results are presented in Section~\ref{sec:specific_instruction}.


\subsection{Analyzing the Influence of Instruction Tuning Data}
\label{sec:instruction_tuning_gain_less}
\subsubsection{Large and Diverse Datasets are NOT Suitable for Lightweight Models}
\label{sec: multi_instruction_data}
In this section, we aim to investigate the impact of instruction tuning on lightweight models. \cite{DBLP:conf/iclr/WeiBZGYLDDL22} proposed that the more tasks or datasets involved in instruction tuning, the greater the improvement in the model's performance. To explore whether the same conclusion applies to lightweight models, we conduct instruction tuning for the MindLLM-1.3B and MindLLM-3B models using three different datasets: MOSS, MingLi and carefully selected subsets from MingLi. The zero-shot results on C-Eval are shown in Table~\ref{tab: chinese_performence}. For the MMLU evaluation, we conduct instruction tuning using MOSS, Tulu and carefully selected subsets from Tulu separately. The zero-shot results on MMLU are shown in Table~\ref{tab: english_performence}. The few-shot results on C-Eval and MMLU are shown in Appendix~\ref{sec:few-shot-results-appendix}.

\begin{table}[h]
\centering
\resizebox{\linewidth}{!}{
\begin{tabular}{clrccccc}
\hline
\bf Model    & \bf Dataset    & \bf Data Size & \bf STEM  & \makecell{\bf Social Sciences} & \bf Humanities & \bf Others & \bf Avg.      \\ \hline
\multirow{4}{*}{\bf MindLLM-1.3B} & -   & -      & 26.1  & 25.6            & 26.2       & 24.7   & 25.7          \\
&MOSS   & 1,074,551 & 26.7  & 26.8            & 25.3       & 24.9   & 26.1           \\
&MingLi & 556,498   & 26.3 & 26.9           & 25.7      & 25.1  & 26.2     \\
&Sub-MingLi  & 50,000    & 26.2 & 26.4           & 26.7      & 26.5  & \textbf{26.4} \\ \hline
\multirow{4}{*}{\bf MindLLM-3B} & -     & -     & 24.0  & 26.4            & 26.6       & 24.2   & 25.0         \\
&Moss      & 1,074,551 & 23.6  & 24.0            & 25.2       & 23.5   & 24.0           \\
&MingLi   & 556,498   & 23.4  & 24.5            & 26.4       & 23.9   & 24.3       \\
&Sub-MingLi    & 50,000    & 26.3  & 25.9            & 25.9       & 25.5   & \textbf{26.0}         \\ \hline
\end{tabular}
}
\caption{The performance of lightweight models after instruction tuning on varying amounts of data on C-Eval. Specially, Sub-MingLi represents the model's instruction tuning on the subset of 50,000 data samples selected from the MingLi dataset.}\label{tab: chinese_performence}
\end{table}


\begin{table}[h]
\centering
\resizebox{\linewidth}{!}{
\begin{tabular}{clrccccc}
\hline
\bf Model & \bf Dataset        & \bf Data size& \bf STEM & \makecell{\bf Social Sciences} & \bf Humanities  & \bf Others & \bf Avg.            \\ \hline
\multirow{4}{*}{\bf MindLLM-1.3B} & -   & -         & 22.2 & 23.8  & 23.9  & 23.1   & 23.3           \\
& MOSS    & 1,074,551 & 21.4 & 21.7 & 24.2  & 23.7   & 22.9           \\
& Tulu & 417,421  & 23.1 & 25.3 & 25.1  & 23.5   & 24.4           \\
& Sub-Tulu & 50,000 & 25.0 & 25.2 & 24.9  & 24.3   & \textbf{24.8}  \\ \hline
\multirow{4}{*}{\bf MindLLM-3B} & -     & - & 24.9 & 24.0 & 25.3   & 24.2   & 24.5           \\
& MOSS     & 1,074,551  & 24.3 & 23.5 & 24.5   & 26.0   & 24.6           \\
& Tulu  & 417,421  & 23.9  & 23.5 & 24.5   & 26.8  & \textbf{24.7}          \\
& Sub-Tulu   & 50,000  & 24.0 & 24.4 & 25.5   & 23.9   & 24.6           \\ \hline
\end{tabular}
}
\caption{The performance of lightweight models after instruction tuning on varying amounts of data on MMLU. Specially,Sub-Tulu represents the model's instruction tuning on the subset of 50,000 data samples selected from the Tulu dataset.}\label{tab: english_performence}
\end{table}

The MOSS dataset, MingLi dataset and Tulu dataset exhibit higher diversity in terms of both data and tasks. Nonetheless, the experimental results indicate that models trained with these instruction tuning datasets does not exhibit notable enhancements in performance, regardless of whether applied to the 1.3B or 3B model configurations. In Section~\ref{sec:loss_methods} of this paper, we propose an approach to construct an tnstruction set using an entropy-based quality filtering strategy. Through this method, we selected 50,000 instances. Although the quantity of instruction tuning data chosen through this approach is notably smaller in comparison to extensive datasets such as MOSS, it possesses the capacity to effectuate enhancements in the model's performance. Regarding the C-Eval metric, the MindLLM model demonstrates performance improvements of up to 1\%, while in the context of the MMLU metric, it exhibits a performance enhancement of up to 1.5\%.


\paragraph{Data Quality is More Important Than Data Diversity and Quantity.} Although the MOSS dataset, MingLi dataset and Tulu dataset exhibit higher data and task diversity, the performance of lightweight models improves more on smaller but higher-quality datasets. Specifically, there is an improvement of 0.7\% and 1\% on C-Eval and 1.5\% and 0.1\% on MMLU for the lightweight models. Therefore, for lightweight models, instruction tuning with higher-quality data is more effective than using larger and more diverse datasets.



\subsubsection{Improve Specific Abilities through Instruction Tuning on Tailored Data}
\label{sec:specific_instruction}
In Section~\ref{sec: multi_instruction_data}, we draw the conclusion that lightweight models cannot achieve emergent abilities on a large and highly diverse dataset. In order to investigate whether lightweight models can be improved in specific abilities through targeted instruction tuning, we conducted experiments on the MindLLM-1.3B model.

\paragraph{Enhance MindLLM-1.3B's proficiency in Chinese subject knowledge.} We use 4,000,000 exam samples in Wanjuan-1.0 dataset, which contains knowledge from subjects such as mathematics, history, biology and others. The results are shown in Table~\ref{tab: exam_instruction}.

\begin{table}[h]
\small
\begin{tabular}{lcccccc}
\hline
\bf Model             & \bf STEM & \bf Social Sciences & \bf Humanities & \bf Others & \bf Avg. \\ \hline
GPT-4\citep{DBLP:journals/corr/abs-2303-08774}              & 67.1 & 77.6            & 64.5       & 67.8   & 68.7     \\
Baichuan-13B~\citep{yang2023baichuan} & 47 & 66.8 & 57.3 & 49.8 & 53.6 \\
InternLM-7B~\citep{2023internlm}        & 48   & 67.4            & 55.4       & 45.8   & 52.8      \\
ChatGLM2-6B~\citep{du2022glm} & 48.6 & 60.5 & 51.3 & 49.8 & 51.7 \\
\hline
Chinese-Alpaca-33B~\citep{chinese-llama-alpaca} & 37   & 51.6            & 42.3       & 40.3   & 41.6  \\
Yuren-13b\footnote{https://github.com/pleisto/yuren-13b
}          & 36.9 & 48.8            & 40.7       & 38.9   & 40.4      \\
ChatGLM-6B~\citep{du2022glm}  & 33.3 & 48.3 & 41.3 & 38 & 38.9 \\
llama-65B~\citep{touvron2023llama} & 37.8 & 45.6 & 36.1 & 37.1 & 38.8 \\
Llama2-Moses-7B~\citep{DBLP:journals/corr/abs-2307-09288}    & 31.2 & 41.4            & 36.2       & 32.4   & 34.5    \\
MOSS~\citep{sun2023moss} & 31.6 & 37 & 33.4 & 32.1 & 33.1 \\\hline
MindLLM-1.3B          & 26.1 & 25.6            & 26.2       & 24.7   & 25.7     \\
\textbf{MindLLM-1.3B*}   & \textbf{41.0} & \textbf{51.8}            & \textbf{41.6}       & \textbf{33.5}   & \textbf{41.6}       \\ \hline
\end{tabular}
 \caption{Comparison of model performance on C-Eval. * represents the MindLLM model fine-tuned using the Wanjuan dataset for instruction tuning. MindLLM-1.3B model achieves a 15.4\% increase in accuracy on C-Eval after targeted instruction tuning.}\label{tab: exam_instruction}
\end{table}


After subject knowledge instruction tuning, we test the model on C-Eval to ascertain whether there is a noticeable improvement in performance in that domain. As shown in Table~\ref{tab: exam_instruction}, the MindLLM-1.3B model achieved a 15.4\% increase in accuracy on C-Eval after targeted instruction tuning. Its performance exceeds that of models with dimensions of 7B, 13B, and even 33B.

\paragraph{Enhance MindLLM-1.3B's proficiency in English reasoning ability.} To augment the reasoning capabilities of MindLLM-1.3B, we perform instruction tuning on the model utilizing the LogiCoT dataset~\citep{DBLP:journals/corr/abs-2305-12147}, which serves as an instruction set for teaching models of logical reasoning and elicits general reasoning skills. 

We evaluate the model's reasoning capabilities as outlined in Section~\ref{sec:complex_reasoning}. The findings demonstrate that, subsequent to instruction tuning with the LogiCoT dataset, the model's accuracy in \textbf{Integrated reasoning} rises from 18.22\% to 23.22\%, reflecting a \textbf{5\%} enhancement. Furthermore, the model's accuracy in \textbf{Knowledge reasoning} improves from 35.76\% to 38.12\%, constituting a \textbf{2.36\%} increase.  

\paragraph{Lightweight models demonstrate outstanding performance in specific domains.} Although lightweight models may not exhibit substantial performance gains on broadly diversified general datasets, they can enhance particular capabilities via instruction tuning using datasets tailored to the target domain. Thus, we believe that lightweight models exhibit significant potential when deployed in domain-specific tasks. In Section~\ref{sec:application}, we show the concrete applications of the MindLLM model within the domains of legal and financial domains.

\subsubsection{Explanation of Performance Decline under Few-shot}
\label{sec:few_shot_explanation}
The inability to achieve substantial performance enhancements when using an extensive and diverse set of instruction tuning data has caught our attention. In an effort to explain this matter, we investigate the performance of the model under both zero-shot and few-shot scenarios, thereby deriving noteworthy insights.

\begin{table}[h]
\small
\begin{tabular}{lcccccc}
\hline
\multicolumn{1}{c}{\multirow{2}{*}{\bf Model}} & \multicolumn{2}{c}{\bf C-Eval}                & \multicolumn{2}{c}{\bf MMLU}         & \multicolumn{2}{c}{\bf CMMLU}        \\
\multicolumn{1}{c}{}                        & zero-shot           & five-shot           & zero-shot           & five-shot  & zero-shot           & four-shot  \\ \hline
MindLLM-1.3B                                   & 25.7                & 26.0                & 23.3                & 26.6       & 25.0                & 25.1       \\
MindLLM-1.3B$\ddagger$                                   & 26.4(+0.7)          & \textbf{27.2(+1.2)} & \textbf{24.8(+1.5)} & 26.9(+0.3) & \textbf{25.2(+0.2)} & 24.7(-0.4) \\ \hline
MindLLM-3B                                     & 25.0                & 25.9                & 24.7                & 26.0       & 24.8                & 24.9       \\
MindLLM-3B$\ddagger$                                     & \textbf{26.0(+1.0)} & 25.8(-0.1)          & \textbf{24.6(-0.1)} & 24.7(-1.3) & \textbf{25.4(+0.6)} & 25.2(+0.3) \\ \hline
\end{tabular}
\caption{Comparing the increase in accuracy in zero-shot and few-shot scenarios after instruction tuning. $\ddagger$ represents the instruction tuning model.}\label{tab: zero_few_shot}
\end{table}

From the results in Table~\ref{tab: zero_few_shot}, it can be observed that, after instruction tuning, the accuracy improvement on zero-shot scenarios for the 1.3B and 3B models on C-Eval, MMLU, and CMMLU dataset is 0.7\%, 1.5\%, 0.2\%, and 1.0\%, -0.1\%, 0.6\%, respectively. However, the accuracy improvement on few-shot scenarios is 1.2\%, 0.3\%, -0.4\%, -0.1\%, -1.3\%, and 0.3\%, respectively, which is lower than the improvement observed in zero-shot scenarios.

The improvement in few-shot scenarios on various datasets is not substantial after instruction tuning, and the accuracy even decreases in some cases. The reason behind this is that instruction tuning disrupts some of the inherent data formats and generation rules learned by the model during the pre-training stage. During the pre-training stage, the model is trained by next-token prediction approach, which is similar to a continuation task. This methodology enables the model to emulate prior instances, thereby equipping it to generate appropriate responses in few-shot tasks. However, during the instruction tuning stage, the model's focus shifts from learning perse to the acquisition of the correlations existing between instructions and corresponding answers. This strategic adjustment enables the model to enhance its ability to identify optimal solution pathways for novel instructions.

Due to the inherent challenges that lightweight models encounter in acquiring the correlations existing between instructions and corresponding answers, the introduction of complex and diverse instructions provides less benefits. Meanwhile, it disrupts their capacity to emulate prior instances  in few-shot scenarios.


\subsection{An Entropy-based Quality Filtering Strategy for Selecting Instruction Set}
\label{sec:loss_methods}
In Section~\ref{sec: multi_instruction_data}, we derive the conclusion that optimization of the model's general performance is more proficiently realized through the amelioration of instruction data quality, as opposed to mere augmentation in data quantity. In the subsequent section, we will introduce an approach to constructing an instruction set using an entropy-based quality filtering strategy. We elucidate the criteria delineating high-quality instruction tuning data for lightweight models and demonstrate that the inclusion of such data contributes to the augmentation of the model's general performance.

\subsubsection{Construct an Instruction Set Using an Entropy-based Quality Filtering Strategy}
The quality of instruction tuning data will directly impact the model's performance~\citep{DBLP:conf/emnlp/WangMAKMNADASPK22}. \cite{DBLP:conf/iclr/WeiBZGYLDDL22} employ the diversity of instruction tuning data as a proxy for measuring the quality of instruction tuning. In our investigation, we determine that employing data entropy and data length constitutes a more appropriate approach to characterizing the quality of instruction tuning data, particularly in the context of lightweight models. In order to filter high-quality data suitable from large-scale instruction tuning data, we propose an approach to construct an instruction set using an entropy-based quality filtering strategy that relies on the prior knowledge of pre-trained models to filter high-quality data. Specifically, we define the cross-entropy loss of the pre-trained model on each instruction data as the \textit{data entropy}. This score serves as an evaluative measure of the model's alignment with the provided data.

We input the Chinese instruction dataset MingLi with 566,498 samples and English instruction dataset Tulu with 417,421 samples (mentioned in Section~\ref{sec:instruction_runing_combination}) into the pret-rained model, which then returns the data entropy for each input sample. Following that, we use the \textbf{K-Means} clustering method~\citep{macqueen1967some} to cluster data under various data entropy.  The specific data and data entropy distribution on MindLLM-1.3B and MindLLM-3B are presented in Table~\ref{tab: cluster} and Appendix~\ref{sec:loss_distribution_3b_appendix} respectively.

\begin{table}[h]
\centering
\setlength\tabcolsep{8pt}
\begin{tabular}{crcrc}
\hline
\multicolumn{1}{c}{\multirow{2}{*}{\bf Cluster Category}} & \multicolumn{2}{c}{\bf MingLi} & \multicolumn{2}{c}{\bf Tulu} \\ \cline{2-5} 
            & \bf Data Size          & \bf Data Entropy           & \bf Data Size          & \bf Data Entropy           \\ \hline
cluster\_1 & 31,915             & 1.791           & 31,137             & 1.325          \\
cluster\_2 & 91,557             & 2.221           & 66,280             & 1.877          \\
cluster\_3 & 104,360            & 2.510           & 89,017             & 2.280          \\
cluster\_4 & 119,499            & 2.761           & 69,790             & 2.676         \\
cluster\_5 & 107,154            & 3.012           & 62,423             & 3.139       \\
cluster\_6 & 73,062             & 3.293         & 56,136             & 3.606          \\
cluster\_7 & 33,098             & 3.666          & 33,512             & 4.149          \\
cluster\_8 & 5,853              & 4.365          & 9,126              & 4.979          \\ \hline
Total       & 556,498            & 2.758          & 417,421            & 2.727          \\ \hline
\end{tabular}
\caption{Data entropy clustering results of MingLi and Tulu dataset on MindLLM-1.3B. Data Entropy represents the average score within this cluster of data.}\label{tab: cluster}
\end{table}

\subsubsection{Performance with Different Data Entropy}
We abandon clusters that do not reach the threshold of 50,000 examples and randomly sample 50,000 examples from each of the remaining clusters for instruction fine-tuning. In this approach, we obtain 5 clusters of data from both MingLi and Tulu datasets based on their data entropy. We evaluate the models' performance trained on different clusters separately on the C-Eval (Chinese) and MMLU (English), and the zero-shot results are shown in Table~\ref{tab: category_chinese} and Table~\ref{tab: category_english}. The five-shot results are shown in Appendix~\ref{sec:five_shot_appendix}.

\begin{table}[h]
\small
\centering
\begin{tabular}{lcccccccc}
\hline
\bf Model                & \bf Cluster    & \bf STEM & \bf Social Sciences & \bf Humanities & \bf Others & \bf Avg.            & \bf Data Entropy \\ \hline
\multirow{7}{*}{MindLLM-1.3B} & w/o sft     & 26.1    & 25.6               & 26.2          & 24.7      & 25.7                & 1.67 \\
                      & MingLi   & 26.3    & 26.9               & 25.7         & 25.1      & 26.2             & 3.37 \\
                      & cluster\_0 & 25.2    & 26.6               & 27.6          & 26.3      & 26.2               & 2.22 \\
                      & \textbf{cluster\_1} & 26.2    & 26.4               & 26.7          & 26.5      & \textbf{26.4}               & 2.51 \\
                      & cluster\_2 & 24.8    & 26.5              & 26.9          & 26.4      & 25.9               & 2.76 \\
                      & cluster\_3 & 25.0    & 26.3               & 27.0          & 26.2      & 25.9      & 3.01 \\
                      & cluster\_4 & 24.7    & 26.4               & 26.9          & 26.3      & 25.8                 & 3.29 \\ \hline                 
\multirow{7}{*}{MindLLM-3B}   & w/o sft     & 24.0 & 26.4            & 26.6       & 24.2   & 25.0               & 1.83 \\
                      & MingLi   & 23.4 & 24.5            & 26.4       & 23.9   & 24.3               & 2.71 \\
                      & cluster\_0 & 22.6 & 23.8            & 24.2       & 23.8   & 23.4               & 1.93 \\
                      & \textbf{cluster\_1} & 26.3 & 25.9            & 25.9       & 25.5   & \textbf{26.0}   & 2.33 \\
                      & cluster\_2 & 24.7 & 24.8            & 25.9       & 24.7   & 24.9             & 2.65 \\
                      & cluster\_3 & 24.2 & 26.1            & 23.9       & 24.1   & 24.5             & 2.95 \\
                      & cluster\_4 & 22.5 & 25.3            & 24.3       & 24.6   & 23.9               & 3.28 \\ \hline
\end{tabular}
\caption{Zero-shot evaluation results of the 1.3B and 3B models after Chinese instruction tuning on C-Eval.}\label{tab: category_chinese}
\end{table}

\begin{table}[h]
\small
\centering
\begin{tabular}{lccccccc}
\hline
\bf Model                & \bf Cluster    & \bf STEM & \bf Social Sciences & \bf Humanities & \bf Others & \bf Avg.  & \bf Data Entropy \\ \hline
\multirow{7}{*}{MindLLM-1.3B} & w/o sft  & 22.2 & 23.8 & 23.9  & 23.1   & 23.3 & 2.02 \\
                      & Tulu   & 23.1 & 25.3 & 25.1   & 23.5 & 24.4 & 3.58 \\
                      & cluster\_0 & 22.2 & 23.1  & 24.5       & 23.5   & 23.5 & 1.88 \\
                      & cluster\_1 & 22.2 & 22.8 & 24.3 & 23.5   & 23.3 & 2.28  \\
                      & cluster\_2 & 24.6 & 24.2   & 24.0       & 22.9   & 23.9 & 2.68 \\
                      & cluster\_3 & 25.0 & 25.2  & 24.9  & 24.3   & \textbf{24.8} & 3.14 \\
                      & cluster\_4 & 25.0 & 23.9  & 25.4       & 23.2   & 24.4 & 3.61 \\ \hline
                      
\multirow{7}{*}{MindLLM-3B}   & w/o sft     & 24.9 & 24.0            & 25.3       & 24.2   & 24.5 & 1.74 \\
                      & Tulu   & 24.5    & 23.5               & 23.9          & 26.8      & \textbf{24.7}    & 1.90 \\
                      & cluster\_0 & 23.3 & 22.7            & 24.1       & 26.2   & 24.1 & 0.83 \\
                      & cluster\_1 & 24.4 & 23.7            & 23.7       & 27.0   & 24.6 & 1.24 \\
                      & cluster\_2 & 24.0 & 24.4            & 25.5       & 23.9   & 24.6 & 1.94 \\
                      & cluster\_3 & 23.8 & 23.4            & 24.5       & 26.8   & 24.6 & 2.36 \\
                      & cluster\_4 & 22.7 & 22.1            & 24.1       & 24.2   & 23.4 & 2.84 \\ \hline
\end{tabular}
\caption{Zero-shot Evaluation results of the 1.3B and 3B models after English instruction tuning on MMLU. }\label{tab: category_english}
\end{table}


The data presented in Table~\ref{tab: category_chinese} and Table~\ref{tab: category_english} reveals variations in MindLLM-1.3B and MindLLM-3B zero-shot results on both C-Eval and MMLU benchmarks across different levels of average data entropy. On the C-Eval dataset, MindLLM-3B exhibits the highest performance improvement of 1\% with a corresponding data entropy of 2.33, while the model's performance decreases by 1.1\% when the data entropy is 3.28 after instruction tuning. Similarly, on the MMLU dataset, MindLLM-1.3B shows a performance improvement of 1.5\% when trained on data with a data entropy of 3.14, but there is no performance improvement when the data entropy is 2.28. Hence, a marked disparity is observed in the overall capability performance of the lightweight model following instruction tuning, as it pertains to various data clusters.

It is noted that the performance of MindLLM-3B on the MMLU metric contradicts our conclusions. The model did not manifest any discernible performance enhancement subsequent to instruction tuning. As we continue with the ongoing training of the MindLLM-3B model, we anticipate delivering more refined and comprehensive results in forthcoming work.


\paragraph{Across different models, varying data entropy levels exhibit similar performance fluctuations.} Following instruction tuning using distinct data clusters, the lightweight model demonstrates diverse outcomes in benchmark evaluations. It is worth noting that, across multiple models and benchmarks, the data clusters exhibiting superior performance consistently reside within a specific data entropy range, and varying data entropy levels exhibit similar performance fluctuations. Recognition of this range can serve as a valuable guide for selecting high-quality instruction tuning data tailored to lightweight models.

\subsubsection{Which Cluster Corresponds to Better  Performance based on Data Entropy?}
\label{sec:different_loss}


\begin{figure}[!t]
    \centering
    \subfloat[Zero-shot results on C-Eval]{
        \includegraphics[width=0.5\textwidth]{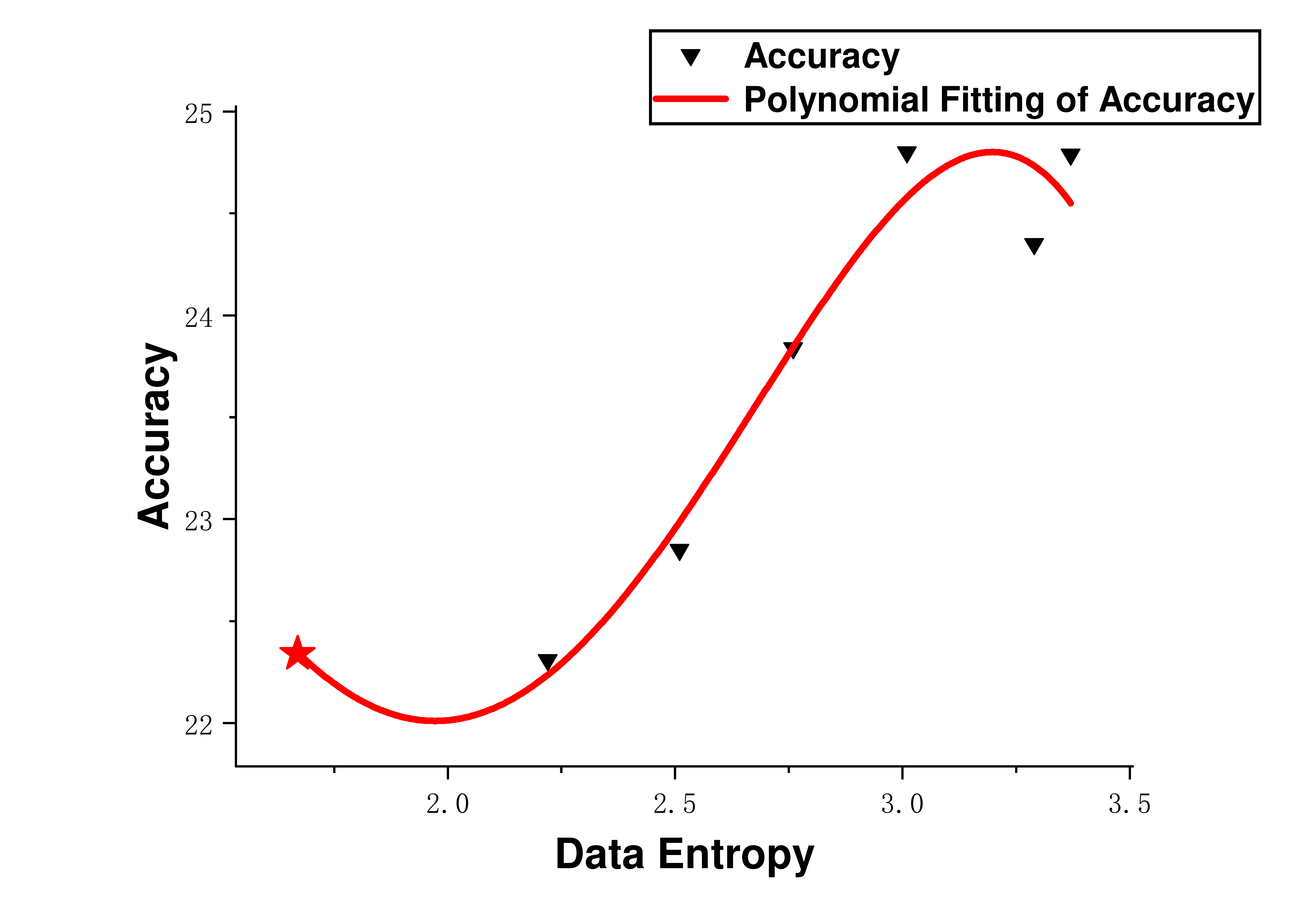}
    }
    \subfloat[Zero-shot results on MMLU]{
        \includegraphics[width=0.5\textwidth]{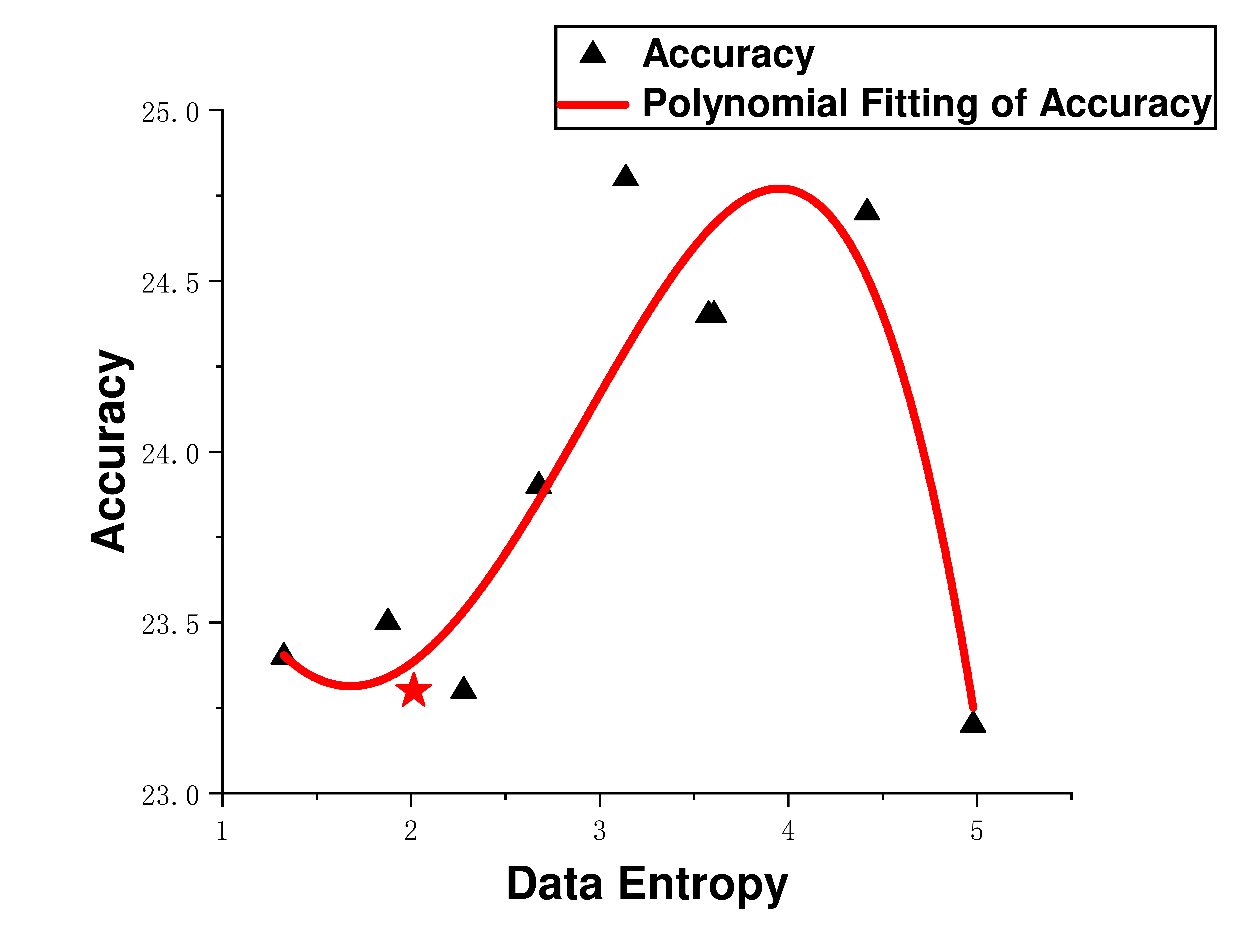}
    }
    \\
    \subfloat[Five-shot results on C-Eval]{
        \includegraphics[width=0.5\textwidth]{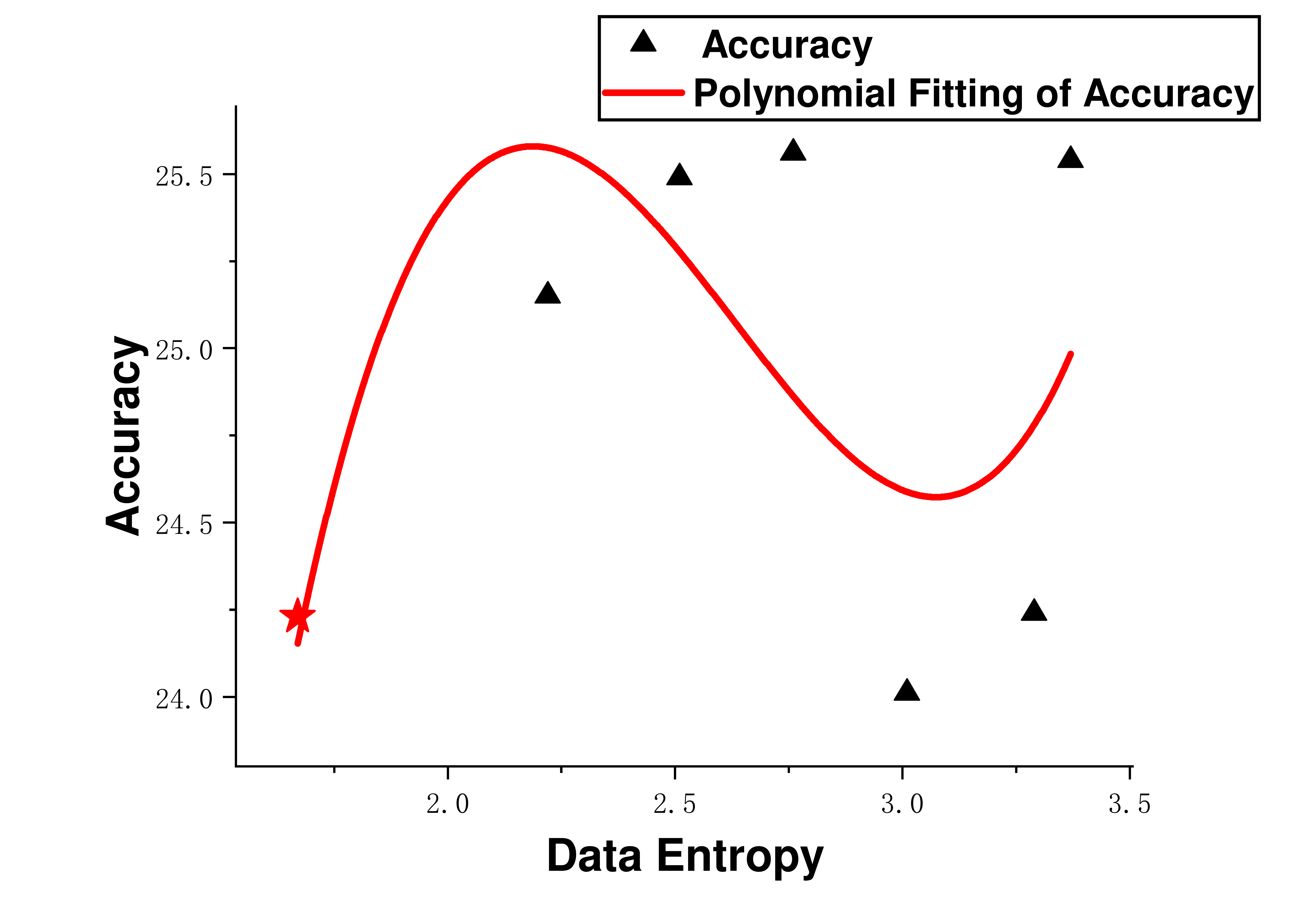}
    }
    \subfloat[Five-shot results on C-MMLU]{
        \includegraphics[width=0.5\textwidth]{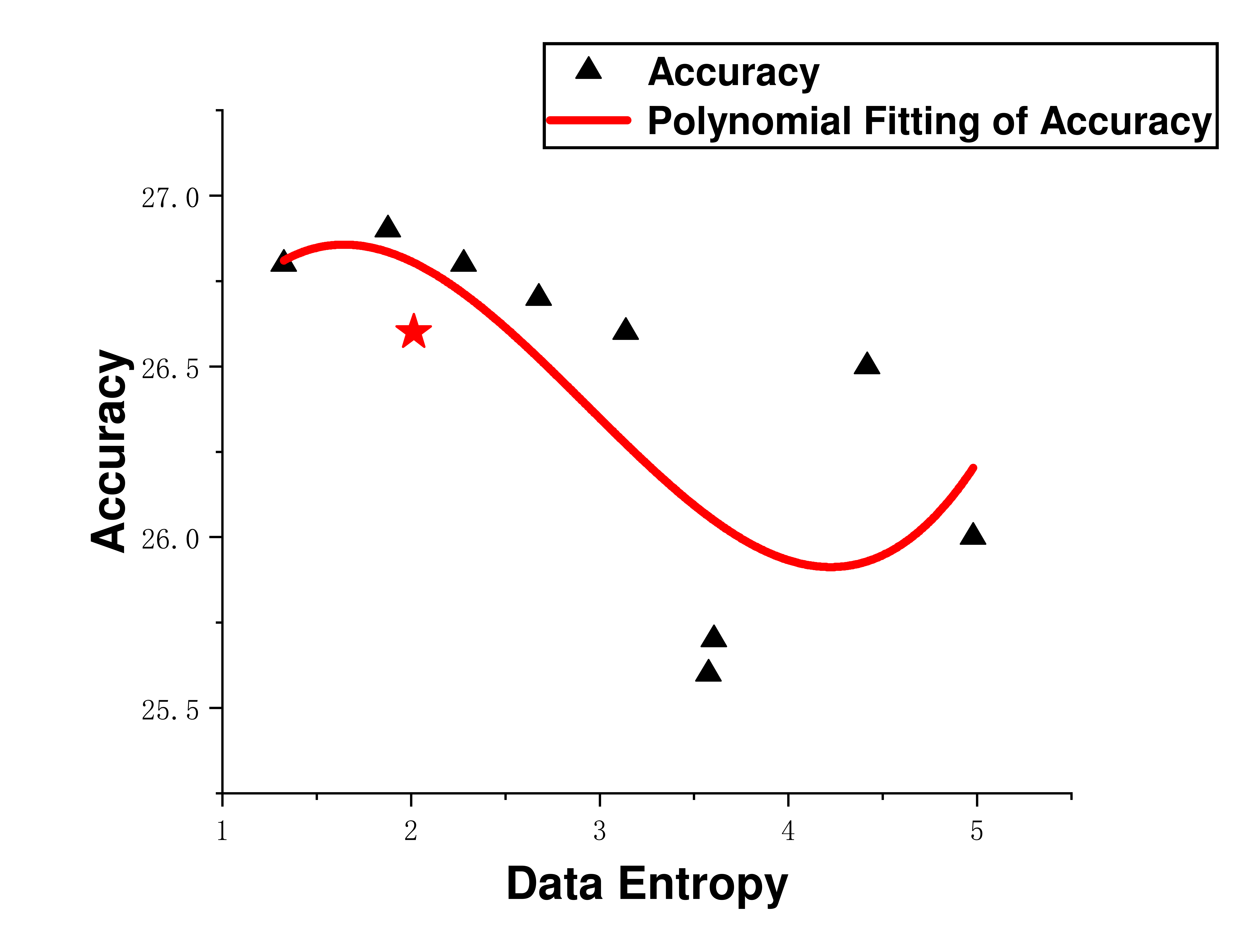}
    }
    \caption{Different data entropy range on the accuracy of the MindLLM-1.3B model in both zero-shot and five-shot scenarios. The {\color{red}$\star$} points represent the pre-training data entropy and its corresponding accuracy of the pre-trained model.}
    \label{fig:loss_figure}
\end{figure}



\paragraph{The trend of model accuracy as data entropy changes} The patterns observed in Figure~\ref{fig:loss_figure}(a) and Figure~\ref{fig:loss_figure}(b) indicate that in the zero-shot scenario, the accuracy of the model exhibits a distinctive pattern. This pattern is characterized by an initial decline, followed by an ascent, and subsequently a further decrement. These variations in accuracy are dependent on the data entropy in proximity to the pre-training entropy (pre-training loss). The models trained on data approximate the pre-training entropy exhibit a trough in accuracy. As for the patterns observed in Figure~\ref{fig:loss_figure}(c) and Figure~\ref{fig:loss_figure}(d) in the five-shot scenario, the models trained on data around the pre-training entropy attain a relatively high level of accuracy. As data entropy expands, accuracy exhibits a gradual descent. Notably, as the data entropy continues to increase beyond that point, a slight upward trend becomes discernible.

Therefore, it becomes evident that the instruction tuning data, which significantly enhances zero-shot performance, does not correspond with the dataset primarily responsible for substantial improvements in five-shot performance. On the one hand, as we conclude in Section~\ref{sec:few_shot_explanation}, it is challenging for lightweight models to comprehensively understand diverse instructions. In the five-shot scenario, the model may rely more on the imitation capabilities learned during the model's pre-training stage to perform downstream tasks. As a result, the requisite instruction tuning data for five-shot scenarios diverges from that which is optimal for zero-shot scenarios. On the other hand, we conduct further analysis in Section~\ref{sec:loss_and_length} to explore the relationship between data entropy, data token count, and model performance, providing a more detailed explanation of this issue. Here, we present our conclusion:
\begin{itemize}
    \item In zero-shot scenarios, selecting data from the range with \textbf{data entropy 1 to 1.5 higher than the pre-training entropy} results in the best performance for lightweight models. Higher or lower data entropy tend to result in a decline in model performance.
    \item In five-shot scenarios, selecting data from the range with \textbf{data entropy 0.5 to 1 higher than the pre-training entropy} results in the best performance for lightweight models. The model's performance is poorest when the data entropy is in the middle range, and there is an improvement in model capabilities as the data entropy continues to increase.
\end{itemize}

\subsubsection{The Relationship Among Model, Data Entropy, and Data Token Count}
\label{sec:loss_and_length}
In Section~\ref{sec:different_loss}, we present two innovative discoveries: 
\begin{itemize}
    \item[1.] The data optimal for the instruction tuning of lightweight models can be methodically chosen within a specific data entropy range.
    \item[2.] The range of data entropy conducive to enhanced model performance in the five-shot scenario is observed to be lower than that which prevails in the zero-shot scenario.
\end{itemize}

In this section, we conclude that the length of instruction tuning data has a discernible impact on the distribution of data entropy, consequently instigating fluctuations in model performance. We conduct a statistical analysis of the correspondence among data entropy, data token count, and model performance, and the results are presented in Table~\ref{tab: chinese_data_length} and Table~\ref{tab: english_data_length}.

\begin{table}[h]
\centering
\begin{tabular}{ccccccccccc}
\hline
\bf Model & \bf Instruction & \bf Output & \bf Total Num. & \bf Data Entropy & \bf Accuracy      \\ \hline
\multirow{5}{*}{MindLLM-1.3B}  & 27.8        & 232.9  & 260.6      & 2.22 & 22.3            \\
                       & 32.3        & 174.2  & 206.5      & 2.51 & 22.9         \\
                       & 30.0        & 174.2  & 204.2      & 2.76 & 23.8           \\
                       & 27.3        & 170.1  & 197.4      & 3.01 & \textbf{24.8} \\
                       & 25.4        & 154.9  & 180.3      & 3.29 & 24.4           \\ \hline
\multirow{5}{*}{MindLLM-3B}    & 25.3        & 247.4  & 272.7      & 1.93 & 23.4   \\
                       & 24.5        & 233.6  & 258.1      & 2.33 & \textbf{26.0} \\
                       & 27.9        & 173.7  & 201.6      & 2.65 & 24.9  \\
                       & 29.7        & 154.1  & 183.8      & 2.95 & 24.5  \\
                       & 31.9        & 126.7  & 158.6      & 3.28 & 23.9     \\ \hline
\end{tabular}
\caption{The relationship between data entropy, data token count, and model performance in the Chinese dataset.}\label{tab: chinese_data_length}
\end{table}

\begin{table}[h]
\centering
\begin{tabular}{cccccc}
\hline
\bf Model & \bf Instruction & \bf Output & \bf Total Num. & \bf Data Entropy & \bf Accuracy      \\ \hline
\multirow{5}{*}{MindLLM-1.3B}  &  78.9        & 208.4  & 287.3      & 1.88 & 23.5          \\
                       & 123.5       & 167.7  & 291.2      & 2.28 & 23.3          \\
                       & 135.5       & 89.5   & 225        & 2.68 & 23.9          \\
                       & 114.0       & 39.9   & 153.9      & 3.14 & \textbf{24.8} \\
                    & 77.7        & 26.5   & 104.2      & 3.61 & 24.4          \\ \hline
\multirow{5}{*}{MindLLM-3B}    & 69.5        & 258.0  & 327.5      & 0.83 & 24.1          \\
                      & 65.8        & 243.8  & 309.6      & 1.24 & \textbf{24.6} \\
                      & 157.3       & 81.3   & 238.6      & 1.94 & \textbf{24.6} \\
                       & 188.1       & 38.8   & 226.9      & 2.36 & \textbf{24.6} \\
                       & 96.9        & 24.3   & 121.2      & 2.84 & 23.4          \\ \hline
\end{tabular}
\caption{The relationship between data data entropy, data token count, and model performance in English dataset.}\label{tab: english_data_length}
\end{table}

According to the results in Table~\ref{tab: chinese_data_length} and Table~\ref{tab: english_data_length}, we observe that as data entropy increases, the token count of the data consistently decreases. This phenomenon is due to the nature of the pre-training stage, wherein the model is engaged in next-token prediction tasks that typically entail longer data sequences. When the data sequences become shorter, a consequent increase in data entropy values is observed.

\paragraph{Explain why the selection strategy with data entropy works.} In the context of zero-shot tasks, it is observed that when the token count resides within the range of 200-250, the model frequently attains a level of performance that can be deemed satisfactory. Excessively large or small token counts tend to adversely affect the model's performance. Our proposition posits that when the data length is excessively long, the instruction tuning data tends to be similar to the pre-training data, and it contains a substantial amount of information, which is not conducive to the lightweight model's instruction tuning. On the other hand, when the data length is excessively short, it contains limited information, and the lightweight model may struggle to acquire significant knowledge from it. Considering that the corresponding data entropy consistently surpasses the pre-training entropy by a margin of 1-1.5 when the token count ranges from 200 to 250, the model excels in performance within this particular entropy range.

\paragraph{Explain the decreased entropy range in the five-shot scenario.} For the five-shot scenario, data typically exhibits lower entropy values when contrasted with the entropy values of the zero-shot scenario. This disparity suggests that the data length is extended in the former compared to the latter. In the five-shot scenario, the input length is longer than in zero-shot scenarios, resembling the format of the continuation tasks learned during the pre-training stage. Therefore, longer data tends to yield better results during instruction tuning in the five-shot scenario and the corresponding data entropy consistently surpasses the pre-training entropy by a margin of 0.5-1.

%



\section{Applications}
\label{sec:application}

To verify that small models can deliver outstanding performance in specific domains, we conduct explorations in the legal and financial fields. 
In specific domains, we observe that the model's parameter size has a limited impact on performance. Our model's performance surpasses that of models of comparable scale and is comparable to larger models.

\subsection{Law}
\subsubsection{Supervised Fine-Tuning}
\paragraph{Dataset and Model}
We collect some publicly available legal-related data, including scenario-based question-answer pairs with legal references from LaW-GPT~\citep{LAWGPT-zh} and a set of NLP-based legal tasks from DISC-LawLLM~\citep{yue2023disclawllm}, which contains a pair instruction dataset and triplet instruction dataset. To ensure that the model retains its innate reasoning abilities and can understand human instructions, we incorporate two general instruction datasets for training, which are generated by GPT-4 with Alpaca prompts translated into Chinese by ChatGPT~\citep{peng2023instruction} and are sampled from Firefly~\citep{Firefly}. We conduct Supervised Fine-Tuning on the above datasets using both the 1.3B and 3B versions of MindLLM.

\paragraph{Preprocessing}
 Given the narrower context window of the lightweight midels, we apply additional length-based filtering to the combined datasets. Since the context length for MindLLM-1.3B is 1024, to ensure a fair comparison with MindLLM-3B, we tokenize the data separately using the tokenizers of both models. We then filter out samples that meet the criterion of having a token length not exceeding 450 tokens for both tokenizers. 

To explore how different token lengths impact the models, we extract data separately for token lengths ranging from 200 to 300 tokens and from 300 to 450 tokens for both MindLLM-1.3B and MindLLM-3B models, ensuring an even dataset size between these length ranges. We initially train both the MindLLM-1.3B and MindLLM-3B models using data with a length not exceeding 450 tokens. Subsequently, we separately train these models using data falling within the 200 to 300 tokens range and the 300 to 450 tokens range. The final dataset statistics and models are presented in the Table~\ref{tab:data_process}.

\begin{table}[h]
    \centering
    \scalebox{0.88}{
    \begin{tabular}{lrrrrrr}
        \hline
        \multirow{2}{*}{\bf Dataset} & \multirow{2}{*}{\bf Total} & \multirow{2}{*}{\bf <450} &\multicolumn{2}{c}{\bf 1.3B Tokenizer} & \multicolumn{2}{c}{\bf 3B Tokenizer} \\
        \cline{4-5} \cline{6-7}
         & &  & \bf 200-300 & \bf 300-450 & \bf 200-300 & \bf 300-450 \\
        \midrule
        LaWGPT-Pairs & 92,373 & 91,324 &  38,269 & 13,107 & 13823 & 359 \\
        DISC-SFT-Pair & 166,758 & 71,128 & 12,791 & 30898 & 20,916 & 16,399 \\
        DISC-SFT-Triplet & 16,000 & 642 & 21 & 620 & 370 & 255 \\
        \hline
        Alpaca\_GPT4\_Zh & 48,818 & 46,945 & 7,683 & 14,086 & 13,374 & 2,672 \\
        Firefly & 60,008 & 60,008 & 6,648 & 8,946 & 8,615 & 2,787 \\
        \hline
        Total & 383,949 & 270,047 & 65,412 & 67,657 & 57,098 & 22,472 \\
        \hdashline
        Training Model & - & \makecell{MindLLM\\1.3B\&3B} & \makecell{MindLLM\\1.3B} & \makecell{MindLLM\\1.3B}  & \makecell{MindLLM\\3B} & \makecell{MindLLM\\3B} \\
        \hline
    \end{tabular}
    }
    \caption{Training datasets and models. To ensure a balanced dataset size for different token lengths, we extract a smaller number of samples from the data for each tokenizer within the two length range. This approach guarantees consistent dataset sizes across various token lengths. Specifically, for the 1.3B Tokenizer, the dataset usage is 40,250 (13,107 + 12,791 + 21 + 7,683 + 6,648), while it is 22,472 for the 3B tokenizer.}
    \label{tab:data_process}
\end{table}

\subsubsection{Evaluation}
We use a language model with a large number of parameters to evaluate the generation of our model and compare it with other models. Specifically, we use ChatGPT as the evaluator for the models' outputs. We first evaluate MindLLMs of different sizes trained on different datasets and subsequently select the best-performing MindLLM to assess it alongside other open-source models.

\paragraph{Dataset}
Given the absence of a publicly available objective benchmark in the legal domain, we use a robust LLM model as an evaluator to assess the inference results of our models. We extract 100 multi-turn legal consultation dialogues generated by ChatGPT based on reference legal statutes from a publicly available dataset~\citep{lawyer-llama-report} for evaluation.

\paragraph{Method}
In our observation, when ChatGPT is tasked with scoring model outputs on a scale of 1 to 5, the resulting score distribution exhibits striking similarity with limited discernibility. This phenomenon persists whether scoring is conducted for individual model outputs or multiple outputs simultaneously. Consequently, we are contemplating the utilization of ChatGPT for result ranking. While inherently more subjective than scoring, this approach has the potential to harness the capabilities of large language models more effectively.
The English version of the prompt we provide to ChatGPT is as follows:
\begin{center}
\fcolorbox{black}{gray!10}{\parbox{0.98\linewidth}{\scriptsize \texttt{Here is a conversation between a customer and a lawyer:\\\{dialog\}\\\\ Now there are multiple responses generated by different models to the first query of the customer:\\\\\{models\_responses\}\\\\Please refer to the conversation between the customer and the lawyer, and provide a combined ranking of the responses from the various models based on the following three aspects, along with a brief explanation of the reasons:\\1. Accuracy: The content and semantics of the model's response should align with the reference conversation and accurately reference relevant legal statutes or regulations.\\2. Completeness: The model's response should not omit any details present in the reference conversation.\\3. Clarity: The model's response should demonstrate a rigorous and clear legal logical analysis, with well-structured and coherent sentences.\\\\Please provide the following format for the ranking and explanations:\\Ranking Results: Model Name > Model Name > Model Name\\Reasons:}}}
\end{center}


where \texttt{\{dialog\}} represents the multi-turn conversation between the customer and the lawyer, and \texttt{\{models\_responses\}} represents responses generated by different models. The prompt requests ChatGPT to provide reasons to offer a more robust and reliable foundation for the ranking results.

\paragraph{Elo Rating System}
The Elo rating system is a mathematical method used to assess the relative skills of participants in competitive games. Taking participants A and B as an example, the scoring update formula for A is as follows:
\begin{gather}
    New\_Score = Old\_Score + 32 * (Game\_Result - Expected\_Result), \\
    Game\_Result = \begin{cases}
            	1, & if A > B \\
            	0, & if B > A	
		          \end{cases}, \\
    Expected\_Result = \frac{1}{1 + 10 ^{(S_B - S_A) / 400}},
\end{gather}
where $S_A$ and $S_B$ represent the scores of A and B, and the initial score is set to 1500.

ChatGPT provides an overall ranking and reasoning for each sample across $n$ models, from which we only extract the ranking information. Since Elo scoring necessitates results from comparison between two participants, we transform the overall rankings of $n$ models into pairwise comparisons for $C(n,2)$ model combinations, where $C(n,2) = n! / (2 * (n-2)!)$. For each model combination, we update the scores separately once.

\subsubsection{Results and Analysis}
We first assess the training results of MindLLM of different sizes on various datasets. After obtaining the best-performing model, we evaluate this model alongside other open-source models, including two general models: ChatGLM2-6B~\citep{du2022glm} and Baichuan2-7B-Chat~\citep{baichuan2023baichuan2}, three models finetuned with our dataset: Bloom~\citep{Bloom}, GPT-Neo~\citep{gpt-neo} and Open-LLaMA~\citep{openlm2023openllama}, as well as three models specifically adapted for the legal domain: DISC-LawLLM~\citep{yue2023disclawllm}, ChatLaw~\citep{cui2023chatlaw}, and Lawyer-LLama~\citep{lawyer-llama-report}. Additionally, we apply SFT to Bloom~\citep{Bloom} and GPT-Neo~\citep{gpt-neo} using the same dataset as the best model from MindLLM.

\begin{minipage}[t]{0.45\textwidth}
\centering
    \begin{tabular}{rrrr}
        \toprule
        \bf \makecell{Model\\Size} & \bf \makecell{Dataset\\Length} & \bf \makecell{Dataset\\Number} & \bf \makecell{Elo\\Score} \\
        \midrule
        3B & <450 & 270k & 1668 \\
        3B & 300-450 & 22k & 1630 \\
        1.3B & 200-300 & 40k & 1572 \\
        1.3B & 300-450 & 40k & 1457 \\
        3B & 200-300 & 22k & 1361 \\
        1.3B & <450 & 270k & 1337 \\

        \bottomrule
    \end{tabular}
    \captionof{table}{Elo rating results for MindLLM of different sizes on various datasets.}
    \label{tab:elo-mind}
\end{minipage}
\begin{minipage}[t]{0.53\textwidth}
\centering
    \begin{tabular}{rrr}
        \toprule
        \bf \makecell{Open-source\\Model} & \bf \makecell{Model\\Size} & \bf \makecell{Elo\\Score} \\
        \midrule
        ChatGLM2 & 6B & 2329 \\
        Lawyer-Llama & 13B & 2153 \\
        ChatLaw & 13B & 2000 \\
        DISC-LawLLM & 13B & 1842 \\
        \bf MindLLM-Law & 3B & 1623 \\
        Baichuan2-Chat & 7B & 1616 \\
        Bloom & 1.7B & 1326 \\
        Bloom & 3B & 1187 \\
        GPT-Neo & 1.3B & 1024 \\ 
        GPT-Neo & 2.7B & 902 \\
        Open-LLaMA & 7B & 623 \\

        \bottomrule
    \end{tabular}
    \captionof{table}{Elo rating results for the comparison of Mind-Law with other open-source models.}
    \label{tab:elo-openmodels}
\end{minipage}

Table~\ref{tab:elo-mind} shows the Elo rating results for MindLLM models of varying sizes trained on datasets with different lengths. Notably, MindLLM-3B, trained on a maximum of 270k data, attains the highest score, whereas MindLLM-1.3B, utilizing the same dataset, ranks the lowest. This observation hints at the possibility that, for models with smaller parameter sizes, an excess of training data could lead to overfitting, resulting in a decline in overall performance.

Furthermore, when considering datasets of varying lengths, MindLLM-3B exhibits superior performance when trained on data falling within the 300 to 450 token range, in contrast to its performance with data ranging from 200 to 300 tokens. Conversely, MindLLM-1.3B shows improved performance with shorter-length datasets. This phenomenon underscores the notion that models with larger parameter sizes tend to adapt more effectively to longer training data, while smaller models excel when dealing with shorter-length datasets, likely due to their narrower context windows.

We designate MindLLM-3B, trained on datasets with a length not exceeding 450 tokens, as our premier legal model, named MindLLM-Law. Subsequently, we apply SFT on Bloom, GPT-Neo and Open-LLaMA using the same dataset. Additionally, we fine-tune Open-LLaMA using the LoRA~\citep{hu2022lora} method. The conclusive comparative results can be found in Table~\ref{tab:elo-openmodels}. 
While it's evident that our model MindLLM-Law hasn't surpassed those with 13B parameters and ChatGLM2-6B, it does outperform Baichuan2-7B-Chat, fine-tuned Open-LLaMA-7B, and models with equivalent or smaller parameter sizes. This highlights the strong domain adaptability of our MindLLM, enabling it to achieve performance comparable to models with larger parameter sizes.


\subsection{Finance}
The finance industry has been greatly impacted by the use of large-scale models. Sentiment analysis has become a crucial tool for revealing opinions expressed in financial articles, news, and social media, shaping market understanding. This paper presents an empirical study that aims to validate the performance of our sentiment classification model, specifically designed for the financial domain. The study utilizes a sentiment classification dataset consisting of financial text data.

\begin{table}[h]
\centering
\begin{tabular}{ccccc}
\hline
              & \textbf{Begin Date} & \textbf{End Data}   & \textbf{Total Num}   & \textbf{Datasize}       \\ \hline
Train Dataset & 2011-05-13 & 2022-12-31 & 508,138     & 320,000        \\
Test Dataset  & 2023-01-01 & 2023-08-31 & 113,247     & 20,000         \\ \hline
\end{tabular}
\caption{Finance Dataset. The data from EastMoney was collected via web crawling spanning the time period from 2011-05-13 to 2023-08-31. The dataset is then partitioned into a training dataset and a test dataset based on 2023-01-01.}
\label{tab:FinanceDataset}
\end{table}

\subsubsection{Dataset} 
Based on the research objectives, EastMoney\footnote{https://www.eastmoney.com/default.html} is selected as the data source for this study. As one of the largest and most influential financial and securities portal websites in China, EastMoney encompasses a wide range of financial information and economic news, encompassing finance, stocks, funds, futures, bonds, foreign exchange, banking, and insurance, across various financial sectors. In order to obtain the necessary data for our research, we follow \cite{DBLP:journals/corr/abs-2306-06031} and conduct web crawling of the EastMoney website's content spanning from May 13, 2011, to January 1, 202. The initial training dataset comprises a total of 508,138 records. Following this, we narrowe down the test dataset to the period between January 1, 2023, and August 31, 2023, containing a total of 113,247 records.

\paragraph{Preprocessing and Training} In this study, we format the financial sentiment dataset into a text classification task, where the input consists of financial news, and the output comprises labels corresponding to five categories: positive, negative, neutral, very negative, and very positive. 

We perform the following data preprocessing steps. Firstly, we utilize the stock price fluctuations as labels for news content, categorizing them into five classes. Secondly, we conduct data sampling from the complete dataset. Due to the imbalance in the classification data, we select the label with the smallest quantity as the reference point and randomly sample an equal number of records from the remaining categories. As a result, we obtain a final training dataset consisting of 320,000 records and a testing dataset containing 20,000 records. 

Finally, we conduct training on the dataset using two distinct training methods, each utilizing specific templates. The first method referred to as \textbf{SFT}, exclusively employs the supervised fine-tuning method with Template 1 for training the financial sentiment classification dataset. In contrast, we introduce a novel distillation method, named \textbf{COT}, in the second method to enhance the capabilities of smaller models. Specifically, we incorporate the extraction of rationales, denoted as \textit{Reason} in Template 2, from large language models, such as ChatGPT, as supplementary information to train our model.

\begin{center}
\fcolorbox{black}{gray!10}{\parbox{0.98\linewidth}{\scriptsize \texttt{
\\
\textbf{Template 1}: \\ \#\#\#~Instruction: What is the sentiment category of the following news? Answer category: \{very negative/negative/neutral/positive/very positive\}.\\ \#\#\#~Input: [Finance News]\\ \#\#\#~Response: [Category]. 
\\ \\ 
\textbf{Template 2}: \\ \#\#\#~Instruction: Please answer the viewpoint of this news article. News: [Finance News] \\ \#\#\#~Input: [Reason]\\ \#\#\#~Response: [Category].
\\
}}}
\end{center}


\begin{table}[]
\centering
\begin{tabular}{lccc|ccc}
\hline
\textbf{Model}            & \makecell{\textbf{Training}\\\textbf{Method}} & \makecell{\textbf{Data}\\\textbf{Size}}       & \textbf{Accuracy} & \makecell{\textbf{Training}\\\textbf{Method}} & \makecell{\textbf{Data}\\\textbf{Size}}       & \textbf{Accuracy} \\ \hline
BaiChuan-7B-chat & -                 & -               & 50.69\% (+31.35\%)         & -                & -                 & 19.34\%  \\
\textbf{MindLLM-1.3B}     & COT               & 25k             & 47.79\%(+27.81\%)         & SFT              & 320k              & 19.98\%  \\
\textbf{MindLLM-3B}       & COT               & 25k             & 46.40\%(+26.28\%)         & SFT              & 320k              & 20.12\%  \\
Bloom-3B         & COT               & 25k             & 45.93\% (+24.14\%)         & SFT              & 320k              & 21.79\%  \\
ChatGLM2-6b      & -                 & -               & 45.79\% (+25.75\%)         & -                & -                 & 20.04\%  \\
Open-LLaMA-3B    & COT               & 25k             & 30.28\% (+10.17\%)         & SFT              & 320k              & 20.11\%  \\
Open-LLaMA-7B    & COT               & 25k             & 28.38\% (+8.38\%)          & SFT              & 320k              & 20.00\%  \\ \hline
\end{tabular}
\caption{Accuracy for the comparison of
MindLLM-1.3B and 3B with other open-source models on Finance Dataset.}
\label{tab:FinanceExperimentResult}
\end{table}

\subsubsection{Results and Analysis}
In this study, to validate the performance of our model, we conduct experiments with other open-source models, such as BaiChuan-7B-chat\citep{DBLP:journals/corr/abs-2309-10305}, ChatGLM2-6b\citep{DBLP:conf/acl/DuQLDQY022}, Open-LLaMA-3B and 7B\citep{openlm2023openllama}, and Bloom-3B~\citep{Bloom}. We employ the identical training method as our model for the Bloom-3B and Open-LLaMA-3B and 7B model, i.e., SFT and COT respectively, while we directly apply the inference abilities of BaiChuan-7B-chat and ChatGLM2-6b, due to their capability of following instructions. Detailed experimental results can be found in Table~\ref{tab:FinanceExperimentResult}.

Although the Bloom3b model shows fairly good performance after SFT, with an accuracy of 21.79\%, the MindLLM-1.3B and 3B models exhibit superior performance over models of similar scale and outperform ChatGLM2-6B and Open-LLaMA-7B following COT, achieving accuracy of 47.79\% and 46.40\% respectively. Furthermore, MindLLM-1.3B and 3B achieve performance competitive with BaiChuan-7B-chat. Moreover, our research results indicate that enhancing MindLLM-1.3B and 3B with supplementary auxiliary information can significantly improve performance. For instance, the utilization of COT training on MindLLM-1.3B and 3B yields a notable enhancement in accuracy, exhibiting an impressive boost of 27.81\% and 26.28\% respectively, in comparison to its SFT counterpart.

\section{Conclusion}

In this paper, we have introduced MindLLM, a novel series of lightweight large language models, featuring 1.3 billion and 3.1 billion parameters. These models have not only proven their competitiveness when compared to existing open-source language models with much larger parameter counts but have also surpassed some of these larger models in performance. We have provided a comprehensive exposition of the methods and techniques employed in the development of our models, with a particular focus on the innovative instruction tuning framework, which has significantly enhanced the specific capabilities of our lightweight models. Furthermore, we have also integrated our models into different domains, showcasing their remarkable flexibility in adapting to specific applications. Our work contributes valuable experience and empirical evidence that underscores the feasibility of creating lightweight large language models for agile adoption within specific domains, an endeavor of great significance for developers and researchers working with limited resources. Ultimately, our work represents a significant step towards lightweight large language models in the era of large language models, offering new insights and opportunities in this field.


\bibliographystyle{unsrtnat}
\bibliography{cas-refs}

\appendix
\section{Evaluation of Pretraining}
\subsection{Values}
\label{sec:evaluation_values}
\paragraph{Evaluation Setting} In this section, we evaluate the model's values in terms of truthfulness, detoxification, and ethics. We implement this evaluation on \textbf{TruthfulQA}(12-shot)~\citep{TruthfulQA}, \textbf{ToxiGen}(6-shot)~\citep{hartvigsen-etal-2022-toxigen}, \textbf{Ethics}(3-shot)~\citep{Ethics} benchmark in lm-evaluation-harness~\citep{lm_harness} framework.
\begin{table}[h]
    \centering
    \begin{tabular}{ccccc}
    \hline
         \textbf{Model}&\textbf{\makecell[c]{Truthfulness}$\uparrow$}&\textbf{\makecell[c]{Toxicity}$\uparrow$}&\textbf{\makecell[c]{Ethics}$\uparrow$}&\textbf{Average$\uparrow$}  \\
    \hline
    BaiChuan-2-7B&30.32&57.13&65.10&50.85\\
    LLaMA-2-7B&32.09 &54.68 &63.31 &50.03\\
    Open-LLaMA-7B&29.15 &56.06 &57.90 &47.70\\
    LLaMA-7B&27.99 &51.6 &62.49 &47.36\\
    GPT-Neo-2.7B&31.87 &51.91 &54.81 &46.20\\
    GPT-Neo-1.3B&31.37 &52.66 &55.79 &46.61\\
    \textbf{MindLLM-1.3B}&33.44 &47.98 &55.66 &45.69\\
    Open-LLaMA-3B&28.30 &50.64 &57.90 &45.61\\
    \textbf{MindLLM-3B}&30.16 &48.94 &54.69 &44.60\\
    Bloom-3B&31.79 &47.77 &53.42 &44.32\\
    Bloom-7B&29.15 &54.89 &48.73 &44.26\\
    GPT-J-6B&28.08 &48.19 &55.64 &43.97\\ \hline
    \end{tabular}
    \caption{results on Values: TruthfulQA(12-shot), ToxiGen(6-shot), Ethics(3-shot)}
    \label{tab:Values}
\end{table}
\subsection{Detailed Results on Benckmarks}
Here, we present the detailed scores of benchmarks ,which are utilized in Section~\ref{sec:pretrain}.
\begin{table}[h]
    \centering
    \begin{tabular}{cccccc}
    \hline
         \textbf{Model}&\textbf{HellaSwag}&\textbf{WinoGrande}&\textbf{PubMedQA}&\textbf{PIQA}&\textbf{MathQA}  \\
    \hline
    MOSS-Base-16B &43.15 &61.32 &66.00 &72.03&27.74\\
    BaiChuan-2-7B &54.14 &70.00 &71.20 &77.80&32.86\\ 
    LLaMA-2-7B  &58.29 &73.80 &70.90 &78.62&29.78\\   
    LLaMA-7B &57.42 &71.35&72.20&79.27&28.61\\
    Open-LLaMA-7B &53.11 &68.03&62.00 &75.57&27.27\\
    Bloom-7B  &46.22 &65.67 &58.90&73.83&26.53\\
    Open-LLaMA-3B &49.56 &64.56 &63.90 &75.62&24.89\\
    Bloom-3B &41.25&57.62 &55.60 &70.18 &25.26\\
     GPT-Neo-2.7B   &42.77&59.98 &54.30 &73.18 &25.39\\
     GPT-Neo-1.3B   &57.06 &38.64 &56.00&70.78&24.49\\
     \textbf{MindLLM3B} &42.40 &58.32  &63.90&71.22&24.89\\
     \textbf{MindLLM-1.3B} &29.93 &50.12 &50.20 &65.07&24.19\\
    \hline
    \end{tabular}
    \caption{Detailed results of complex reasoning: BBH and LogicQA is shown in Table~\ref{tab:Complex_Reasoning}}
    \label{tab:details_of_complex_reasoning}
\end{table}



\section{Instruction Tuning Appendix}
\subsection{Few-shot Results after MindLLM Instruction Tuning}
\label{sec:few-shot-results-appendix}
We separately evaluate the few-shot results of MindLLM-1.3B and MindLLM-3B on CEval and MMLU. Here, we provide the average evaluation results of the models on each benchmark. Here, we conduct experiments using a five-shot scenario, and the results show in Table~\ref{tab: few_shot_performence_appendix}.
\begin{table}[h]
\centering
\begin{tabular}{clccc}
\hline
\bf Models & \bf Dataset       & \bf Data Size & \bf CEval-avg     & \bf MMLU-avg      \\ \hline
\multirow{5}{*}{MindLLM-1.3B} & - & -         & 24.2          & 26.6          \\
& MOSS & 1,074,551 & 22.7          & 23.0          \\
& MingLi & 556,498   & 24.2          & -             \\
& Tulu & 417,421   & -             & 26.6          \\
& Sub &  50,000    & \textbf{25.6} & \textbf{26.9} \\ \hline
\multirow{5}{*}{MindLLM-3B}   & -         & 25.9          & \textbf{26.0} \\
& MOSS & 1,074,551 & 24.7          & 24.1          \\
& MingLi & 556,498   & 25.8          & -             \\
& Tulu & 417,421   & -             & 24.6          \\
& Sub & 50,000    & \textbf{25.9} & 24.7          \\ \hline
\end{tabular}
\caption{The few-shot performance of MindLLM-3B models after instruction tuning on varying amounts of data on CEval and MMLU. Specially, Sub represents the model's instruction tuning on the subset of 50,000 data samples selected from the MingLi or Tulu dataset.}\label{tab: few_shot_performence_appendix}
\end{table}


In few-shot scenarios, large-scale and highly diverse instruction tuning data still do not exhibit outstanding performance, reaffirming our conclusion mentioned in Section~\ref{sec: multi_instruction_data}: For lightweight models, data quality is more important than data diversity and quantity.

\subsection{Specific Data Distribution and Data Entropy Situations on MindLLM-3B Model}
\label{sec:loss_distribution_3b_appendix}

The specific data distribution and data entropy situations on MindLLM-3B are presented in Table~\ref{tab: 3B_cluster_appendix}.

\begin{table}[h]
\centering
\setlength\tabcolsep{8pt}
\begin{tabular}{lcccc}
\hline
\multicolumn{1}{c}{\multirow{2}{*}{\bf Cluster Category}} & \multicolumn{2}{c}{\bf MingLi} & \multicolumn{2}{c}{\bf Tulu} \\ \cline{2-5} 
            & \bf Data Size          & \bf Data Entropy           & \bf Data Size          & \bf Data Entropy           \\ \hline
cluster\_1 & 24,024             & 1.345          & 55,721             & 0.826          \\
cluster\_2 & 59,159            & 1.933          & 102,289             & 1.240          \\
cluster\_3 & 115,939             & 2.327          & 116,865              & 1.577          \\
cluster\_4 & 121,498            & 2.647          & 94,823            & 1.935          \\
cluster\_5 & 120,084           & 2.950          & 69,717             & 2.356          \\
cluster\_6 & 86,901              & 3.282          & 57,148             & 2.844          \\
cluster\_7 & 40,269             & 3.717          & 39,377             & 3.360          \\
cluster\_8 & 7,382             & 4.511          & 8,648             & 4.165          \\ \hline
total       & 556,498            & 2.713          & 544,588            & 1.902          \\ \hline
\end{tabular}
\caption{Data entropy clustering results of MingLi and Tulu dataset on MindLLM-3B. The Data Entropy in the table represents the average score within this cluster of data.}\label{tab: 3B_cluster_appendix}
\end{table}

\subsection{The Model Performances of Different Data Entropy Range in Five-shot}
\label{sec:five_shot_appendix}

We obtain clusters of data from both MingLi and Tulu datasets based on their data entropy. In this context, we furnish insights into the diverse model performance  contingent on the utilization of distinct instruction tuning data in a five-shot scenario. The results are shown in Table~\ref{tab: 1.3_english_five_shot_appendix}.

\begin{table}[h]
\centering
\begin{tabular}{lccccc}
\hline
\bf Dataset  & \bf Data Size & \bf CEval-avg  & \bf Data Entropy   & \bf MMLU-avg & \bf Data Entropy   \\ \hline
w/o sft & -   & 24.2 & 1.67 & 26.6   & 2.02       \\
MingLi & 556,498   & 25.5    & 3.37  & -   & -          \\
Tulu & 417,421   & -       &  -  & 25.6  & 3.58        \\
cluster\_0 & 50,000   & 25.2 & 2.22 & \textbf{26.9} & \textbf{1.88} \\ 
cluster\_1 & 50,000   & 25.5 & 2.51  & 26.8 & 2.28 \\ 
cluster\_2 & 50,000   & \textbf{25.7} & \textbf{2.76} & 26.7 & 2.68 \\ 
cluster\_3 & 50,000  & 24.0 & 3.01 & 26.6 & 3.14 \\ 
cluster\_4 & 50,000  & 24.4 & 3.29 & 25.7 & 3.61 \\ \hline
\end{tabular}
\caption{Evaluation five-shot results of the MindLLM-1.3B after instruction tuning on C-Eval and MMLU.}\label{tab: 1.3_english_five_shot_appendix}
\end{table}

We conclude that in five-shot scenarios, selecting data from the interval with data entropy 0.5 to 1 higher than the pretraining entropy results in the best performance for lightweight models. The model’s performance is poorest when the data entropy is in the middle interval, and there is an improvement in model capabilities as the data entropy continues to increase.

\subsection{Different Data Entropy Ranges on The Accuracy of MindLLM-3B Model}
\label{sec:3B_loss_figure_appendix}

\begin{figure}[h]
    \centering
    \subfloat[Zero-shot results on C-Eval]{
        \includegraphics[width=0.5\textwidth]{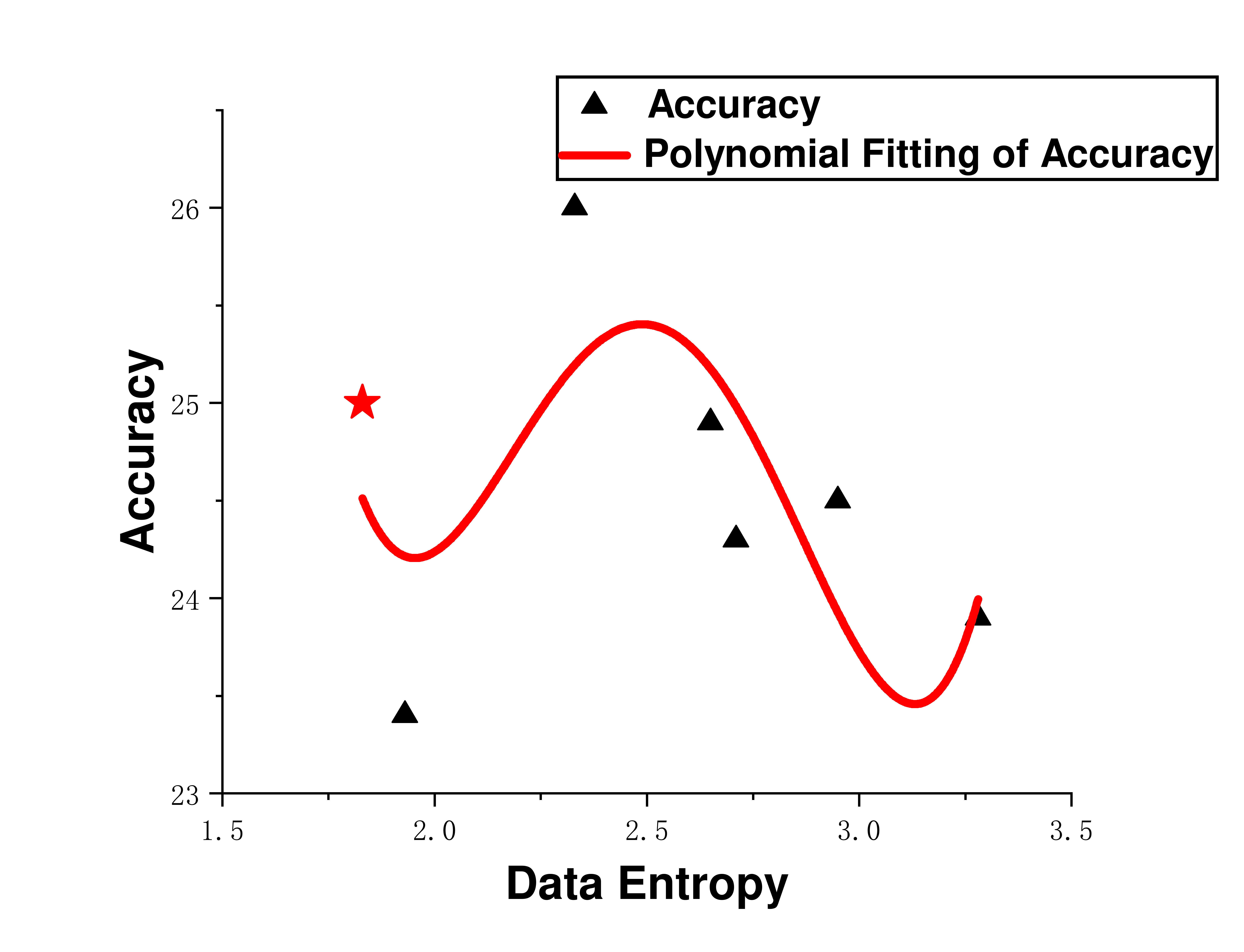}
    }
    \subfloat[Zero-shot results on MMLU]{
        \includegraphics[width=0.5\textwidth]{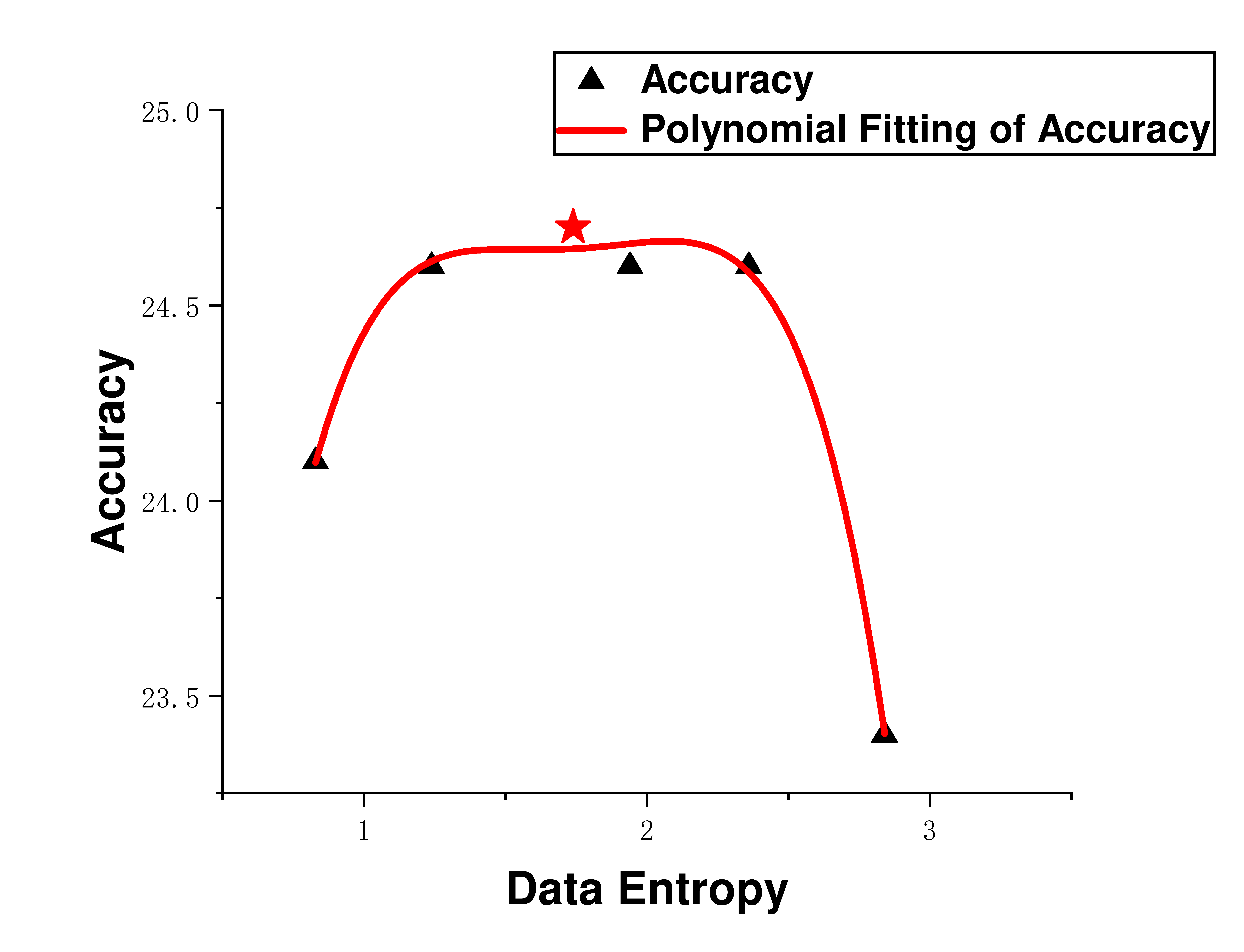}
    }
    \caption{Different data entropy ranges on the accuracy of the MindLLM-3B model in zero-shot scenario. And the {\color{red}$\star$} points represent the pretraining data entropy and its corresponding accuracy of the pretrained model.}
    \label{fig:3B_loss_figure_appendix}
\end{figure}

Figure~\ref{fig:3B_loss_figure_appendix} shows the accuracy of the MindLLM-3B model in zero-shot scenario of different data entropy. It can be observed that he performance of the MindLLM-3B model remains remarkably stable across various instruction tuning processes, rendering it challenging to assess the influence of data with different Data Entropy on MindLLM-3B. In light of the ongoing, uninterrupted training regimen being applied to the MindLLM-3B model, we anticipate furnishing more precise and detailed results in forthcoming work.

\end{document}